\documentclass[lettersize,journal]{IEEEtran}

\usepackage{amsmath,amssymb,amsfonts}
\usepackage{textcomp}

\usepackage{graphicx}
\usepackage{float}
\usepackage{stfloats}
\usepackage{placeins}
\usepackage{adjustbox}
\usepackage{booktabs}
\usepackage{multirow}
\usepackage{makecell}
\usepackage{tabularx}
\usepackage{array}
\usepackage{siunitx}

\usepackage[table,xcdraw]{xcolor}

\definecolor{bestrow}{HTML}{E5E3F1}
\definecolor{secondbest}{HTML}{FAEBD8}
\newcolumntype{Y}{>{\centering\arraybackslash}X}

\usepackage{pifont}
\usepackage{bbding}
\usepackage{verbatim}
\usepackage{lipsum}

\usepackage{algorithm}
\usepackage{algpseudocode}

\definecolor{notegreen}{RGB}{165,195,188}
\algrenewcommand\algorithmiccomment[1]{%
  \hfill{\color{notegreen}$\triangleright$~#1}%
}

\usepackage{caption}
\usepackage{subcaption}
\usepackage{hyperref}
\usepackage{cite}
\usepackage{url}
\usepackage{cleveref}
\usepackage{tabularx}
\usepackage{array}

\newcolumntype{Y}{>{\centering\arraybackslash}X}

\usepackage{pifont}
\usepackage{bbding}
\usepackage{verbatim}
\usepackage{lipsum}
\usepackage{booktabs}
\usepackage[table]{xcolor}
\definecolor{oursrow}{HTML}{EAF7E6}
\newcommand{\cmark}{\ding{51}}  
\newcommand{\xmark}{\ding{55}}  

\begin{document}

\title{PhysVR: Vision-Language Model Guided Interference-aware Temporal Feature Refinement for Remote Physiological Measurement}

\author{Zixu Li, Jianjun Qian,~\IEEEmembership{Member,~IEEE}, Hang Shao, Daoheng Li, Lei Luo, and Jian Yang
\thanks{Zixu Li, Jianjun Qian, Daoheng Li, Lei Luo, and Jian Yang are with the PCA Lab, Key Lab of Intelligent Perception and Systems for High-Dimensional Information of Ministry of Education, School of Computer Science and Engineering, Nanjing University of Science and Technology, Nanjing 210094, China. Hang Shao is with the College of Computer Science and Technology, Qingdao University, Qingdao 266071, China.
(email:lizixu@njust.edu.cn; csjqian@njust.edu.cn; shaohang@qdu.edu.cn; lidaoheng@njust.edu.cn; cslluo@njust.edu.cn; csjyang@njust.edu.cn).

Copyright © 20xx IEEE. Personal use of this material is permitted. However, permission to use this material for any other purposes must be obtained from the IEEE by sending an email to pubs-permissions@ieee.org.
}

}

\markboth{Journal of \LaTeX\ Class Files,~Vol.~14, No.~8, August~2021}%
{Shell \MakeLowercase{\textit{et al.}}: A Sample Article Using IEEEtran.cls for IEEE Journals}

\IEEEpubid{}

\maketitle

\begin{abstract}
Remote photoplethysmography (rPPG) enables contactless physiological measurement from facial videos, yet its subtle pulse-related variations are easily affected by illumination variation, head motion, facial blur, and region-of-interest instability. Existing methods mainly suppress interference during feature learning,
while whether the learned temporal features remain affected by interference
and how to further suppress such interference before rPPG estimation are
rarely examined. To address this limitation, we propose PhysVR, a vision-language model guided interference-aware temporal feature refinement framework for rPPG estimation. Specifically, a physiological backbone produces global temporal features and a coarse rPPG prediction, from which signal-derived physiological reliability evidence is constructed from local temporal characteristics. In parallel, a frozen vision-language model processes sampled facial frames under an interference-oriented prompt, and an evidence head extracts visual interference evidence from the VLM output. Temporal cross-attention integrates the physiological and visual evidence with the global temporal features to construct interference-aware temporal context. Guided by this context, a shared temporal correction unit performs general refinement, while four interference-specific experts selectively suppress different interference through adaptive routing. The refined temporal features are then used for final rPPG estimation. Extensive experiments on five public benchmarks demonstrate that PhysVR consistently outperforms representative methods under both intra-dataset and cross-dataset evaluation protocols.

\end{abstract}

\begin{IEEEkeywords}
Remote photoplethysmography, vision-language models, facial videos
\end{IEEEkeywords}


\begin{figure}[h]
  \centering
  \includegraphics[width=0.985\linewidth]{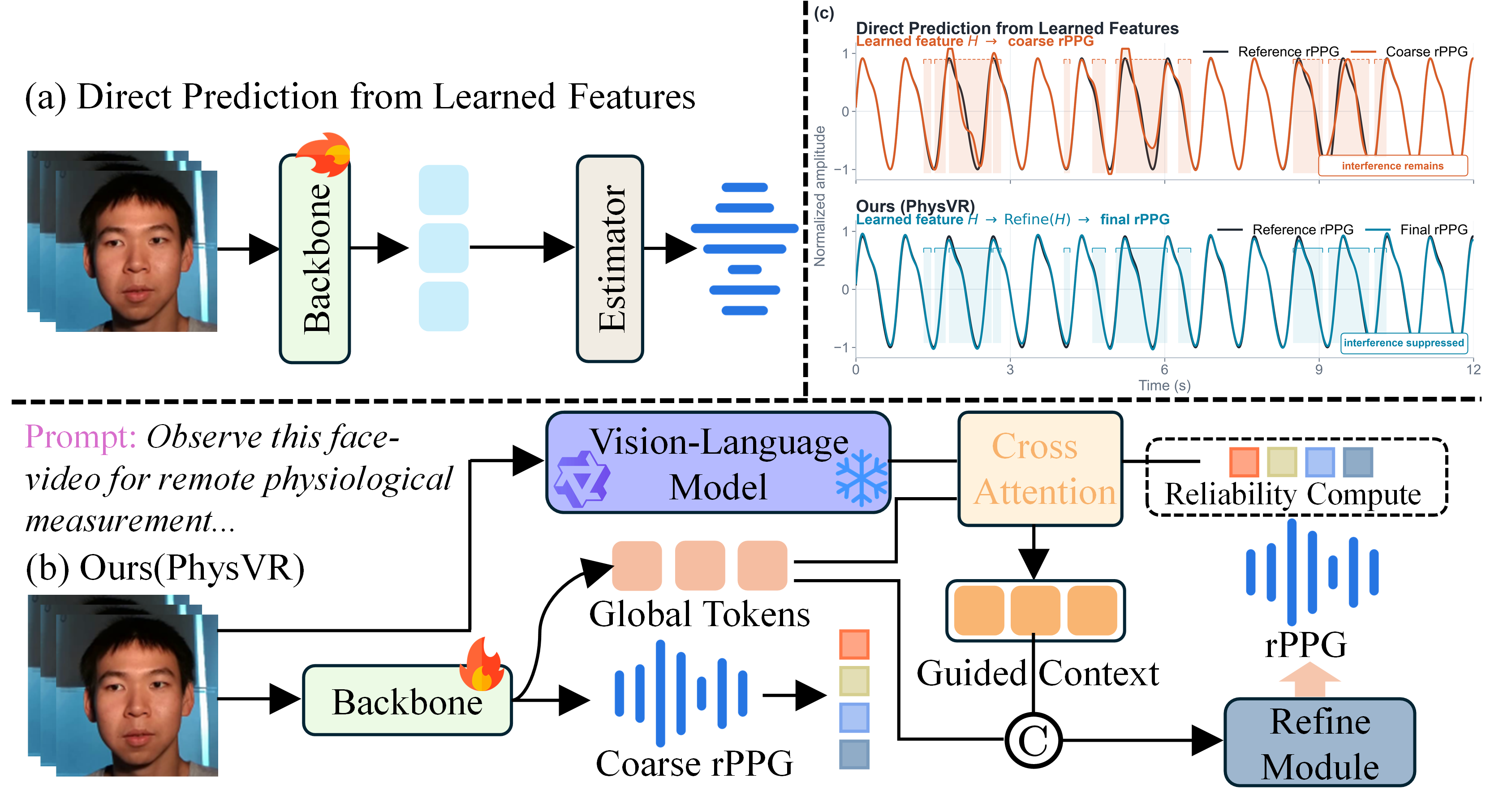}
\caption{Comparison between direct rPPG estimation and the proposed interference-aware temporal feature refinement.
(a) Direct prediction maps learned temporal features directly to rPPG signal.
(b) PhysVR integrates physiological reliability and VLM-derived visual interference evidence to refine temporal features before final estimation.
(c) An example showing that learned temporal features may remain affected by interference under challenging visual conditions, while further refinement improves the final rPPG prediction.}
     \label{fig:fig1}

\end{figure}
\section{Introduction}
\IEEEPARstart
Remote photoplethysmography (rPPG) enables contactless physiological measurement by capturing subtle variations in facial skin color induced by cardiac blood-volume changes. Requiring only a camera and no physical contact, rPPG has been widely investigated for remote physiological monitoring and clinical care~\cite{69,62,liphysflow,pei1}, affective-state analysis~\cite{71}, and mobile physiological sensing~\cite{72}. However, reliable rPPG estimation remains challenging in practical environments, where weak pulse-related variations can be easily obscured by illumination variation, head motion, facial blur, and region-of-interest (ROI) instability.

Existing deep-learning based rPPG methods mainly follow two feature modeling strategies. STMap-based approaches organize temporal color traces extracted from multiple facial regions into structured spatio-temporal maps, whereas video-based approaches learn spatial and temporal features directly from facial clips~\cite{2,DeepPhys}. Their architectures have evolved from convolutional networks to transformer-based and state-space models~\cite{44,Rhythmmamba}. Despite differences in feature extraction and temporal modeling, most existing methods directly regress rPPG signals from the learned features.

Under this direct estimation paradigm, the feature extractor must preserve weak pulse-related variations while suppressing non-physiological changes. Illumination variation alters facial color and intensity distributions, head motion and ROI instability disrupt temporal correspondence across frames, and facial blur attenuates subtle skin-color variations~\cite{3,4,60,61}. Existing methods generally address these
factors during feature learning and subsequently use the learned temporal
features for signal estimation. However, whether these features remain
affected by interference after feature learning and how to further suppress
such interference before prediction are rarely examined. 

Recently, pretrained vision-language models (VLMs) and large language models (LLMs) have been introduced into remote physiological measurement. Existing studies exploit cross-modal knowledge to improve physiological feature learning and signal estimation~\cite{vlphys,Physllm}. rPPG-VQA further combines signal-quality analysis with multimodal scene assessment to select suitable videos for unsupervised rPPG training~\cite{VQArPPG}. These studies demonstrate the potential of language-guided visual understanding for remote physiological measurement. Nevertheless, the potential of VLMs to characterize visual interference for subsequent temporal feature refinement remains largely
unexplored.

To address this problem, we propose PhysVR, a vision-language model guided
interference-aware temporal feature refinement framework for robust rPPG
estimation. Since interference may remain in the learned temporal features
after feature learning, PhysVR further refines these features before final
rPPG estimation. As illustrated in Fig.~\ref{fig:fig1}, most existing methods
directly estimate rPPG from the learned temporal features after feature
learning. However, under challenging visual conditions, these features may
still be affected by interference, leading to deviations in the estimated
rPPG signal. PhysVR therefore introduces further temporal feature refinement
before final estimation, guided by physiological reliability evidence and
visual interference evidence. Specifically, the physiological backbone extracts global temporal features and predicts a coarse rPPG signal. Based on the local temporal characteristics
of the coarse prediction, time-resolved physiological reliability evidence is
derived to reflect its reliability at different temporal positions. In
parallel, a frozen VLM processes sampled facial frames under an
interference-oriented prompt, and an evidence head derives clip-level visual
interference evidence to characterize the visual interference present in the
input video. Temporal cross-attention integrates the physiological reliability
evidence and visual interference evidence with the global temporal features
to construct an interference-aware temporal context. Guided by this context,
a shared temporal correction unit first performs general refinement of the
learned temporal features, while four interference-specific experts further
suppress different types of interference through adaptive routing. The refined
temporal features are then used for final rPPG estimation.

Our contributions are summarized as follows:
\begin{itemize}
\item We propose PhysVR, a vision-language model guided
interference-aware temporal feature refinement framework that further
refines learned temporal features after feature learning, reducing the influence of interference that may remain under challenging visual conditions.

\item We construct an interference-aware temporal context by integrating
physiological reliability evidence and VLM-derived visual interference
evidence with global temporal features through temporal cross-attention,
allowing complementary physiological and visual information to jointly
guide the subsequent refinement process.

    \item We design a context-guided refinement module consisting of a shared
    temporal correction unit, four interference-specific experts, and an
    adaptive temporal router. The shared unit performs general temporal
    refinement, while the router selectively activates interference-specific
    experts to further suppress interference. Experiments on five public benchmarks demonstrate the effectiveness of PhysVR under both intra-dataset and cross-dataset.
\end{itemize}

\section{Related Work}
\label{sec:related_work}

\subsection{Remote Physiological Measurement}

Early rPPG methods relied on handcrafted color-space projections and signal-processing priors to recover pulse signals while suppressing illumination changes and motion artifacts~\cite{chrom,63}. With the development of deep learning, the field gradually shifted toward data-driven spatio-temporal feature learning. Representative architectures include convolutional attention networks~\cite{DeepPhys}, 3D spatio-temporal networks~\cite{64}, temporal-shift models~\cite{TS-CAN}, transformer-based models~\cite{44}, and efficient end-to-end networks~\cite{Efficientphys}. More recent studies have further explored periodic sparse attention, state-space modeling, diffusion-based modeling, and physics-grounded temporal modeling~\cite{Rhythmformer,Rhythmmamba,PhysDiff,PHASE-Net}.

Beyond architectural advances, considerable efforts have focused on improving robustness to environmental interference and domain variation. ND-DeeprPPG disentangles environmental noise from pulse-related features by using background regions as a reference~\cite{13}. Shao et al.~\cite{1} further model complex time-varying interference under real-world and extreme illumination conditions through interference sharing and disentanglement. FLOW improves cross-domain generalization by aligning multi-source features through optimal-transport-driven feature warping~\cite{flow}. Despite these
advances, existing methods mainly suppress or disentangle interference during
feature learning, or reduce distribution discrepancies across domains. The
learned temporal features are then typically used directly for rPPG estimation,
without further interference-aware and refinement before signal prediction.

\subsection{Vision-Language Models}

Vision-language models (VLMs) jointly model visual and textual information to support open-vocabulary recognition, visual grounding, image captioning, and multimodal reasoning~\cite{65,66,67,68}. Their prompt-conditioned features have also been increasingly adapted to domain-specific visual understanding tasks.

Pretrained multimodal models have recently been introduced into remote physiological measurement for different purposes. Yue et al.~\cite{vlphys} align visual and textual features carrying physiological frequency information to facilitate self-supervised rPPG learning. PhysLLM incorporates a large language model (LLM) into rPPG estimation through cross-modal alignment and task-oriented physiological cues~\cite{Physllm}. rPPG-VQA employs a multimodal large language model (MLLM) to jointly assess signal quality and recording conditions, enabling the selection of suitable videos for unsupervised rPPG training~\cite{VQArPPG}. Collectively, these studies exploit pretrained multimodal models for
physiological feature learning, signal estimation, or training-data
selection. However, they primarily use multimodal information to support
physiological representation learning or data assessment, rather than to
provide visual interference information for further refinement of learned
temporal features.

\begin{figure*}[t]
    \centering
    \includegraphics[width=0.985\linewidth]{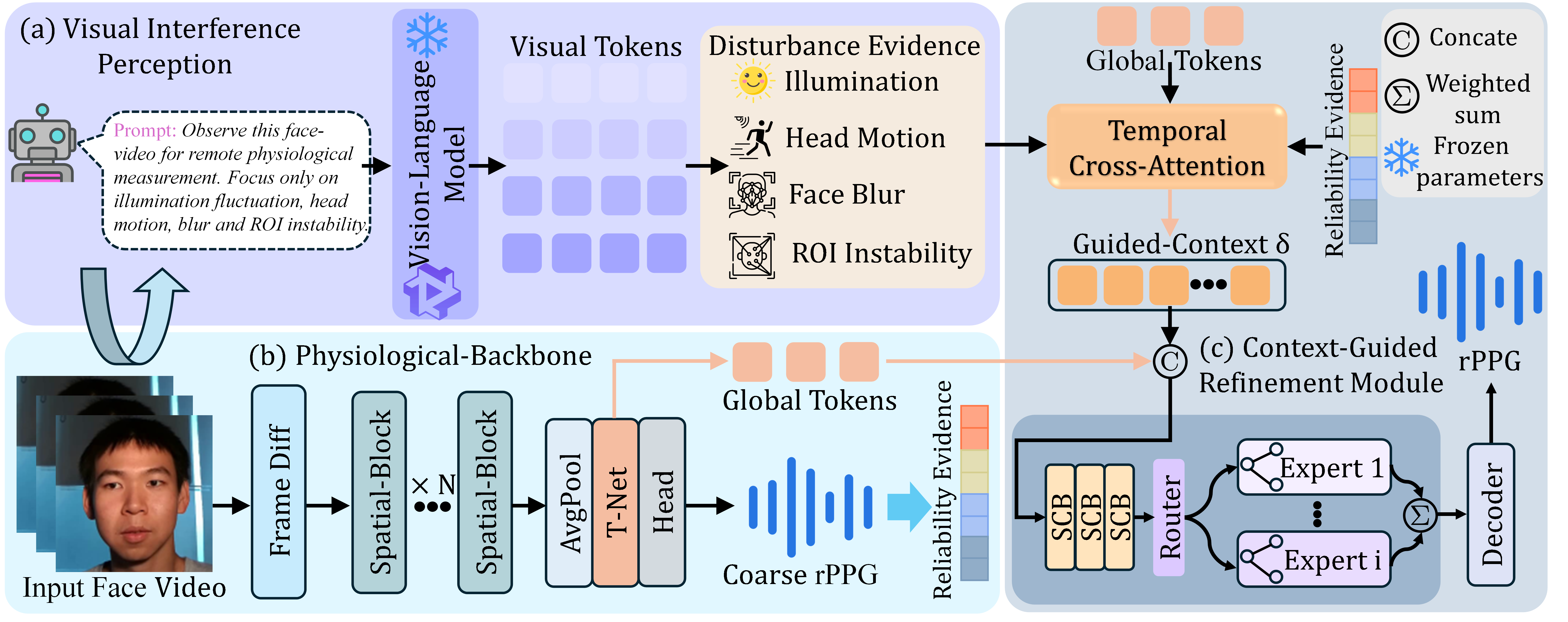}
\caption{Overview of PhysVR. The physiological backbone extracts global
temporal features and derives time-resolved physiological reliability
evidence from a coarse rPPG signal. In parallel, a frozen VLM processes
sampled frames under an interference-oriented prompt, whose outputs are
used to derive clip-level visual interference evidence. Temporal cross-attention
integrates both evidence sources with the global temporal features to
construct an interference-aware temporal context. Guided by this context,
a shared temporal correction unit performs general refinement, while four
interference-specific experts selectively suppress interference through
adaptive routing before final rPPG estimation.}
    \label{Pic_Overview}
\end{figure*}

\section{Methodology}

As illustrated in Fig.~\ref{Pic_Overview}, PhysVR consists of three
main components: a physiological backbone, a VLM-based visual
interference perception branch, and a context-guided refinement module.
The physiological backbone provides the global temporal representation
for rPPG estimation and derives time-resolved physiological reliability
evidence from a coarse prediction, reflecting local variations in
prediction reliability. In parallel, the VLM-based branch characterizes
visual interference present in the input video and produces clip-level
visual interference evidence. These two sources of evidence provide
complementary physiological and visual information for subsequent
temporal feature refinement. The context-guided refinement module then
combines shared temporal refinement with interference-specific experts
and adaptive temporal routing to further refine the learned temporal
features before final rPPG estimation.

\subsection{Physiological Backbone}
\label{sec:backbone}

The physiological backbone extracts global temporal features
$\mathbf{H}$ from frame differences and predicts a coarse rPPG signal
$\hat{\mathbf{s}}^{c}$. The coarse signal is used
to derive time-resolved physiological reliability evidence $\mathbf{R}$.

Given a facial video
$\mathbf{V}=(\mathbf{I}_{1},\ldots,\mathbf{I}_{T})
\in\mathbb{R}^{T\times3\times H\times W}$,
we construct a frame-difference sequence:
\begin{equation}
\begin{aligned}
\Delta\mathbf{V}
&=
\mathcal{D}(\mathbf{V}) \\
&=
\left[
\mathbf{0},
\mathbf{I}_{2}-\mathbf{I}_{1},
\ldots,
\mathbf{I}_{T}-\mathbf{I}_{T-1}
\right].
\end{aligned}
\label{eq:frame_difference}
\end{equation}
The zero tensor $\mathbf{0}$ has the same dimensions as a video frame
and is inserted at the first position to preserve the sequence length.
Frame differencing suppresses temporally invariant facial appearance
while emphasizing adjacent-frame variations, including subtle pulse-related
changes and non-physiological fluctuations caused by visual interference.

\begin{figure}[t]
  \centering
  \setlength{\fboxsep}{0pt}
  \includegraphics[width=\linewidth]{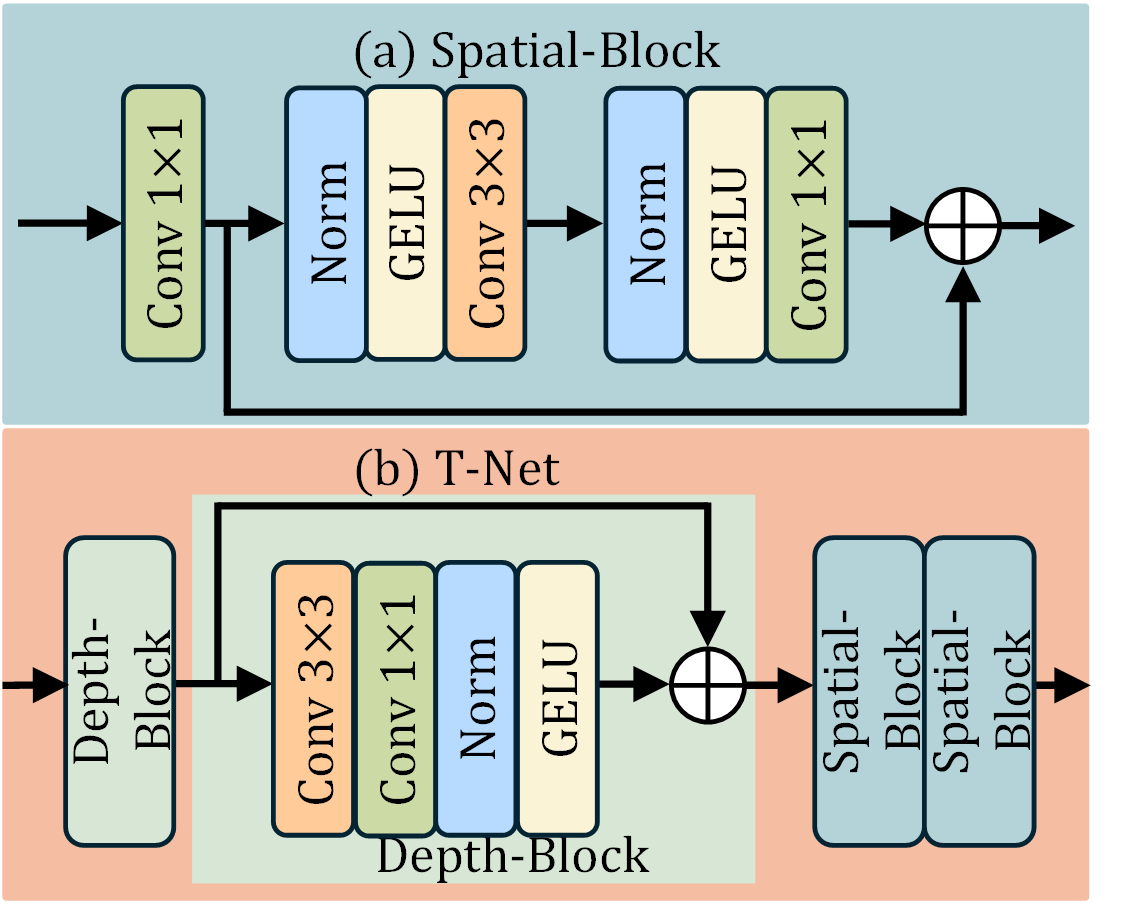}
  \caption{Principal components of the physiological backbone.
  (a) A Spatial-Block uses a residual bottleneck composed of
  pointwise and spatial convolutions. (b) The T-Net contains two Depth-Blocks followed by two Spatial-Blocks.}
  \label{fig:pic_2}
\end{figure}

Spatial features are first extracted from the difference sequence and
then processed along the temporal dimension:
\begin{equation}
\begin{aligned}
\mathbf{F}
&=
E_s\left(
\Delta\mathbf{V}
\right), \\
\mathbf{H}
&=
E_T\left(
E_p\left(
\operatorname{AvgPool}_{\mathrm{sp}}(\mathbf{F})
\right)
\right).
\end{aligned}
\label{eq:temporal_feature_extraction}
\end{equation}
The spatial encoder $E_s(\cdot)$ consists of stacked Spatial-Blocks.
As illustrated in Fig.~\ref{fig:pic_2}(a), each Spatial-Block uses a
residual bottleneck composed of pointwise and spatial convolutions.
Spatial average pooling reduces the spatial resolution, while the
projection layer $E_p(\cdot)$ adjusts the feature dimension for subsequent
temporal modeling. As shown in Fig.~\ref{fig:pic_2}(b), the $E_T(\cdot)$ contains two
Depth-Blocks followed by two Spatial-Blocks to model temporal dependencies.
The output $\mathbf{H}\in\mathbb{R}^{T\times d}$ denotes the global temporal features.

A signal prediction head $h_s(\cdot)$ maps the global temporal features
$\mathbf{H}$ to a coarse rPPG signal:
\begin{equation}
\hat{\mathbf{s}}^{c}
= h_{s}(\mathbf{H}),
\label{eq:coarse_signal}
\end{equation}
where $\hat{\mathbf{s}}^{c}\in\mathbb{R}^{T}$ denotes the coarse rPPG
signal.

We construct four time-aligned descriptors to capture local temporal
characteristics of the coarse rPPG signal:
\begin{equation}
\begin{aligned}
\mathbf{g}_{\mathrm{diff}}
&=
\left|
\nabla_t\hat{\mathbf{s}}^{c}
\right|, \\
\mathbf{g}_{\mathrm{band}}
&=
\mathcal{M}\left(
\left|
\mathcal{H}_{\mathrm{bp}}
(\hat{\mathbf{s}}^{c})
\right|
\right), \\
\mathbf{g}_{\mathrm{drift}}
&=
\left|
\mathcal{H}_{\mathrm{lp}}
(\hat{\mathbf{s}}^{c})
\right|, \\
\mathbf{g}_{\mathrm{dev}}
&=
\left[
1-\sigma\left(h_q(\mathbf{H})\right)
\right]
\odot
\left|
\hat{\mathbf{s}}^{c}
-
\mathcal{M}(\hat{\mathbf{s}}^{c})
\right|.
\end{aligned}
\label{eq:reliability_descriptors}
\end{equation}
The operator $\nabla_t$ denotes the first-order temporal difference,
with a zero inserted at the first position to preserve the sequence
length, and $\mathbf{g}_{\mathrm{diff}}$ measures changes between
adjacent signal values. The fixed band-pass operator
$\mathcal{H}_{\mathrm{bp}}(\cdot)$ retains frequencies within
$0.7$--$3.0$ Hz, while the fixed low-pass operator
$\mathcal{H}_{\mathrm{lp}}(\cdot)$ retains frequencies below
$0.5$ Hz. Both operators are implemented as fixed, differentiable
frequency-domain masks. The operator $\mathcal{M}(\cdot)$ denotes a
nine-frame moving average with boundary padding.
Accordingly, $\mathbf{g}_{\mathrm{band}}$ describes the local magnitude
of the physiological-band component, while
$\mathbf{g}_{\mathrm{drift}}$ captures slowly varying components.
For $\mathbf{g}_{\mathrm{dev}}$, $h_q(\cdot)$ maps
$\mathbf{H}$ to position-wise modulation coefficients through a sigmoid,
which adaptively weight the deviation of the coarse signal from its
local temporal mean.

The coarse rPPG signal and the four descriptors are concatenated along
the feature dimension:
\begin{equation}
\mathbf{G}_{r}
=
\operatorname{Concate}
\left(
\hat{\mathbf{s}}^{c},
\mathbf{g}_{\mathrm{diff}},
\mathbf{g}_{\mathrm{band}},
\mathbf{g}_{\mathrm{drift}},
\mathbf{g}_{\mathrm{dev}}
\right).
\label{eq:reliability_descriptor}
\end{equation}
A one-dimensional temporal convolution maps the concatenated descriptors
to the feature dimension of $\mathbf{H}$:
\begin{equation}
\mathbf{R}
=
h_{\mathrm{rel}}
\left(
\mathbf{G}_{r}
\right).
\label{eq:physiological_reliability}
\end{equation}
The resulting sequence $\mathbf{R}\in\mathbb{R}^{T\times d}$ summarizes
local temporal characteristics related to the reliability of the coarse
prediction and serves as time-resolved physiological reliability evidence.

\subsection{VLM-Based Visual Interference Perception}
\label{sec:vlm}

While $\mathbf{R}$ captures time-resolved signal characteristics related
to coarse-prediction reliability, it does not explicitly characterize
visual interference in the input clip. As shown in
Fig.~\ref{Pic_Overview}(a), the VLM-based branch therefore extracts
clip-level visual interference information from sampled facial frames.
A fixed interference-oriented prompt guides the VLM to focus on visual
factors relevant to rPPG estimation, including illumination fluctuation,
head motion, facial blur, and ROI instability. The same prompt is used
for all clips and datasets.

To cover the visual conditions throughout the clip while limiting the
computational cost of VLM processing, we uniformly sample $N$ frames:
\begin{equation}
\begin{gathered}
\mathcal{S}_{N}(\mathbf{V})
=
\left(
\mathbf{I}_{\tau_1},
\ldots,
\mathbf{I}_{\tau_N}
\right),\\[-1pt]
\tau_i
=
1+
\left\lfloor
\frac{(i-1)(T-1)}{N-1}
\right\rfloor ,
\end{gathered}
\label{eq:frame_sampling}
\end{equation}
where $2\leq N\leq T$. The first and last frames are always selected,
and the remaining sampling positions are distributed across the clip
while preserving the original temporal order.

The sampled frames and the fixed prompt $p$ are jointly processed by
the frozen VLM:
\begin{equation}
\begin{aligned}
\mathbf{Y}_{v}
&=
\Phi_{\mathrm{VLM}}
\left(
\mathcal{S}_{N}(\mathbf{V}),
p
\right),\\
\mathbf{z}_{v}
&=
\mathbf{Y}_{v,\ell^{\star}} .
\end{aligned}
\label{eq:vlm_features}
\end{equation}
We instantiate $\Phi_{\mathrm{VLM}}$ with
Qwen2-VL-2B-Instruct~\cite{qwen2} and keep its parameters frozen
throughout training. The output
$\mathbf{Y}_{v}\in\mathbb{R}^{L_v\times d_v}$ contains the
final-layer hidden states of the multimodal input sequence, where
$L_v$ denotes the valid sequence length and $\ell^{\star}$ denotes
the index of the final non-padding token. The corresponding hidden
state $\mathbf{z}_{v}\in\mathbb{R}^{d_v}$ is directly used as the
VLM feature for subsequent visual evidence extraction.

A trainable evidence head maps $\mathbf{z}_{v}$ to the visual
interference evidence:
\begin{equation}
\mathbf{e}_{v}
=
\tanh\left(
\mathbf{W}_{e,2}
\delta\left(
\mathbf{W}_{e,1}
\operatorname{LN}(\mathbf{z}_{v})
+
\mathbf{b}_{e,1}
\right)
+
\mathbf{b}_{e,2}
\right).
\label{eq:visual_evidence}
\end{equation}
Here, $\mathbf{W}_{e,1}$ and $\mathbf{W}_{e,2}$ are learnable weight
matrices, while $\mathbf{b}_{e,1}$ and $\mathbf{b}_{e,2}$ are the
corresponding bias vectors. $\operatorname{LN}(\cdot)$ denotes layer
normalization, and $\delta(\cdot)$ denotes the GELU activation. The
vector $\mathbf{e}_{v}\in(-1,1)^{d_e}$serves as clip-level visual interference evidence that reflects the visual interference present in the input clip.

\subsection{Context-Guided Temporal Feature Refinement}
\label{sec:refinement}

As shown in Fig.~\ref{Pic_Overview}(c), PhysVR integrates the
physiological reliability evidence $\mathbf{R}$ and visual interference
evidence $\mathbf{e}_{v}$ with the global temporal features $\mathbf{H}$
through temporal cross-attention. Since $\mathbf{R}$ is time-resolved
while $\mathbf{e}_{v}$ is clip-level, $\mathbf{H}$ is used as the query,
and the two evidence sources are projected into the key and value spaces
to construct an interference-aware temporal context for feature refinement.
For $\xi\in\{K,V\}$, the context construction is formulated as:
\begin{equation}
\begin{aligned}
\mathbf{E}_{\xi}
&=
\operatorname{Concate}
\left(
\mathbf{R}\mathbf{W}_{r}^{\xi},
\mathbf{e}_{v}^{\top}\mathbf{W}_{e}^{\xi}
\right),\\
\mathbf{C}
&=
\operatorname{MHA}
\left(
\mathbf{H}\mathbf{W}_{q},
\mathbf{E}_{K},
\mathbf{E}_{V}
\right)
\mathbf{W}_{o}.
\end{aligned}
\label{eq:temporal_context}
\end{equation}

Here, $\mathbf{W}_{q},\mathbf{W}_{r}^{\xi}\in\mathbb{R}^{d\times d_a}$,
$\mathbf{W}_{e}^{\xi}\in\mathbb{R}^{d_e\times d_a}$, and
$\mathbf{W}_{o}\in\mathbb{R}^{d_a\times d}$ are learnable projection
matrices. The key and value sequences
$\mathbf{E}_{K},\mathbf{E}_{V}\in\mathbb{R}^{(T+1)\times d_a}$
consist of $T$ physiological evidence vectors and one clip-level visual
evidence vector. Cross-attention uses the projected global temporal
features as queries to obtain the interference-aware temporal context
$\mathbf{C}\in\mathbb{R}^{T\times d}$.

The temporal context is concatenated with the global temporal features
and projected to dimension $d$. A shared temporal correction unit then
performs general refinement:
\begin{equation}
\begin{aligned}
\boldsymbol{\Delta}_{\mathrm{shr}}
&=
R_{\mathrm{shr}}
\left(
\operatorname{Concate}
\left(
\mathbf{H},
\mathbf{C}
\right)
\right),\\
\mathbf{U}
&=
\mathbf{H}
+
\lambda_{\mathrm{shr}}
\tanh
\left(
\boldsymbol{\Delta}_{\mathrm{shr}}
\right).
\end{aligned}
\label{eq:shared_correction}
\end{equation}

The shared temporal correction unit
$R_{\mathrm{shr}}(\cdot)$ first projects the concatenated features
from $2d$ to $d$, followed by three residual Shared Correction Blocks
(SCBs) and a pointwise projection. Each SCB contains layer normalization,
a depthwise temporal convolution, and two pointwise projections with
GELU activation. The temporal convolutions use a kernel size of $5$
and dilation rates of $1$, $2$, and $4$, respectively, enabling
multi-scale temporal correction. The resulting correction
$\boldsymbol{\Delta}_{\mathrm{shr}}\in\mathbb{R}^{T\times d}$ is
scaled and added to $\mathbf{H}$ to obtain
$\mathbf{U}\in\mathbb{R}^{T\times d}$.

The shared unit captures temporal deviations common to different
interference conditions. To further suppress specific types of interference, we introduce four experts for illumination variation, head
motion, facial blur, and ROI instability:
\begin{equation}
\mathcal{G}
=
\{
\mathrm{ill},
\mathrm{mot},
\mathrm{blur},
\mathrm{roi}
\}.
\label{eq:expert_set}
\end{equation}

A temporal router independently controls the activation of the four
experts at each temporal position:
\begin{equation}
\begin{aligned}
\mathbf{P}
&=
\sigma
\left(
h_{\mathrm{rt}}(\mathbf{U})
\right),\\
\mathbf{M}_{\mathrm{rt}}
&=
\operatorname{Gate}(\mathbf{P})\\
&=
\left[
\mathbf{m}_{\mathrm{ill}},
\mathbf{m}_{\mathrm{mot}},
\mathbf{m}_{\mathrm{blur}},
\mathbf{m}_{\mathrm{roi}}
\right].
\end{aligned}
\label{eq:expert_gate}
\end{equation}

Here, $\mathbf{P}\in(0,1)^{T\times4}$ contains the position-wise
activation probabilities, and each $\mathbf{m}_{g}\in\{0,1\}^{T}$
denotes the binary temporal gate of expert $g$. The four expert gates
are independently predicted, allowing multiple experts to be activated
simultaneously when different interference factors coexist. The binary
gates use a threshold of $0.5$, with a straight-through estimator
adopted during training.

Each expert is conditioned on the visual interference evidence
$\mathbf{e}_{v}$ through expert-specific channel-wise modulation:
\begin{equation}
\begin{gathered}
\boldsymbol{\gamma}_{g},
\boldsymbol{\beta}_{g}
=
G_{g}(\mathbf{e}_{v}),\\
\mathbf{U}_{g}
=
\operatorname{LN}(\mathbf{U})
\odot
\left(
1+
\frac{1}{2}
\tanh(\boldsymbol{\gamma}_{g})
\right)
+
\tanh(\boldsymbol{\beta}_{g}),\\
\boldsymbol{\Delta}_{g}
=
\lambda_{g}
\tanh
\left(
R_{g}(\mathbf{U}_{g})
\right),
\qquad
g\in\mathcal{G}.
\end{gathered}
\label{eq:expert_correction}
\end{equation}
Here, $\boldsymbol{\gamma}_{g},\boldsymbol{\beta}_{g}\in\mathbb{R}^{d}$
are expert-specific modulation parameters applied across the temporal
dimension. The four experts use different transformations for different types of
interference. The illumination expert performs channel-wise correction
using pointwise projections. The motion expert combines first-order
temporal differences with local temporal convolution to capture abrupt
temporal variations. The blur expert models local temporal deviations
from neighboring features to compensate for locally weakened temporal
variations. The ROI-instability expert employs dilated temporal
convolutions with different dilation rates to capture longer-range
temporal inconsistency. Each expert produces an interference-specific
correction $\boldsymbol{\Delta}_{g}\in\mathbb{R}^{T\times d}$.

The activated expert corrections are combined with
$\mathbf{U}$:
\begin{equation}
\begin{aligned}
\widetilde{\mathbf{H}}
&=
\mathbf{U}
+
\sum_{g\in\mathcal{G}}
\mathbf{m}_{g}
\odot
\boldsymbol{\Delta}_{g},\\
\hat{\mathbf{s}}
&=
D_{\theta}
\left(
\widetilde{\mathbf{H}}
\right).
\end{aligned}
\label{eq:final_refinement_and_prediction}
\end{equation}

The decoder $D_{\theta}(\cdot)$ maps the refined temporal features
$\widetilde{\mathbf{H}}\in\mathbb{R}^{T\times d}$ to the final rPPG
signal $\hat{\mathbf{s}}\in\mathbb{R}^{T}$.

\subsection{Training Objective}
\label{sec:objective}

PhysVR produces a coarse rPPG signal $\hat{\mathbf{s}}^{c}$ and a final
prediction $\hat{\mathbf{s}}$. Since the physiological reliability
evidence $\mathbf{R}$ is derived from the coarse prediction,
$\hat{\mathbf{s}}^{c}$ is explicitly supervised to preserve meaningful
temporal information.

For a prediction $\mathbf{y}$ and ground-truth signal $\mathbf{s}$,
the signal loss is defined as:
\begin{equation}
\begin{aligned}
\mathcal{L}_{\mathrm{sig}}(\mathbf{y},\mathbf{s})
={}&
\lambda_{\mathrm{mse}}
\mathcal{L}_{\mathrm{mse}}
+
\lambda_{\mathrm{np}}
\mathcal{L}_{\mathrm{np}}
+
\lambda_{\mathrm{freq}}
\mathcal{L}_{\mathrm{freq}},
\end{aligned}
\label{eq:signal_loss}
\end{equation}
where $\mathcal{L}_{\mathrm{mse}}$ denotes the point-wise MSE loss,
$\mathcal{L}_{\mathrm{np}}=1-\rho(\mathbf{y},\mathbf{s})$ is the
Pearson correlation loss, and $\mathcal{L}_{\mathrm{freq}}$ is the
Smooth L1 distance between the normalized power spectra within the
physiological frequency range.

The overall objective is:
\begin{equation}
\mathcal{L}
=
\mathcal{L}_{\mathrm{sig}}
\left(
\hat{\mathbf{s}},
\mathbf{s}
\right)
+
\lambda_{\mathrm{c}}
\mathcal{L}_{\mathrm{sig}}
\left(
\hat{\mathbf{s}}^{c},
\mathbf{s}
\right).
\label{eq:overall_objective}
\end{equation}
Where $\lambda_{\mathrm{c}}$ controls the coarse-signal supervision.
All trainable components are optimized jointly.
\section{Experiment}
\label{sec:exp}
We evaluate PhysVR on five public rPPG benchmark datasets:  BUAA-MIHR~\cite{39}, VIPL-HR~\cite{37,38}, NIRP-DRV~\cite{40-drv}, NIRP-IND~\cite{40-indoor}, and MMPD~\cite{MMPD}. Both intra-dataset and cross-dataset evaluations are conducted to assess HR estimation accuracy. We first describe the datasets, evaluation metrics, and implementation details, and then compare PhysVR with state-of-the-art methods through quantitative and qualitative analyses. We further analyze the trade-off between estimation accuracy and  computational complexity, followed by ablation studies of the principal components of PhysVR.


\subsection{Datasets}

\textbf{BUAA-MIHR:}
BUAA-MIHR~\cite{39} is designed for rPPG evaluation under low-illumination conditions. It contains 165 RGB facial videos from 15 subjects recorded at 11 illumination levels ranging from 1.0 to 100.0 lux. Each 60-s video was captured at 30 fps with synchronized finger-clip PPG recorded at 60 Hz. We use recordings with illumination intensities of at least 6.3 lux.

\textbf{VIPL-HR:}
VIPL-HR~\cite{37,38} contains 3,130 facial videos from 107 subjects under less-constrained conditions, including variations in illumination, head motion, and acquisition devices. Each video lasts approximately 30 seconds, with synchronized finger BVP signals provided as reference.

\textbf{NIRP-DRV:}
NIRP-DRV~\cite{40-drv} contains 190 RGB and near-infrared facial recordings from 19 subjects collected in realistic driving environments. The recordings include diverse in-vehicle illumination conditions and natural head movements, with synchronized pulse-oximeter measurements as reference.

\textbf{NIRP-IND:}
NIRP-IND~\cite{40-indoor} provides a complementary indoor setting with 15 RGB and near-infrared recordings from 8 subjects. The recordings include both stationary and motion conditions, with synchronized pulse-oximeter measurements as reference.

\textbf{MMPD:}
MMPD~\cite{MMPD} is a mobile rPPG dataset containing 660 one-minute RGB facial videos from 33 subjects. The recordings cover Fitzpatrick skin types III--VI, four illumination conditions, and four activities: stationary, head rotation, talking, and walking. Synchronized finger PPG signals are provided as reference.

\subsection{Performance Metrics}

Following the standard evaluation protocol~\cite{46}, we use three
commonly adopted metrics for heart rate estimation: Mean Absolute Error
(MAE), Root Mean Square Error (RMSE), and Pearson's Correlation
Coefficient ($\rho$). Given $N$ test samples, let $\hat{h}_{i}$ and
$h_{i}$ denote the estimated and ground-truth heart rates of the
$i$-th sample, respectively. MAE and RMSE are defined as:
\begin{equation}
\begin{aligned}
\mathrm{MAE}
&=
\frac{1}{N}
\sum_{i=1}^{N}
\left|
\hat{h}_{i}-h_{i}
\right|,\\
\mathrm{RMSE}
&=
\sqrt{
\frac{1}{N}
\sum_{i=1}^{N}
\left(
\hat{h}_{i}-h_{i}
\right)^{2}
}.
\end{aligned}
\label{eq:hr_error_metrics}
\end{equation}

Pearson's correlation coefficient is defined as:
\begin{equation}
\rho
=
\frac{
\sum_{i=1}^{N}
\left(
\hat{h}_{i}-\bar{\hat{h}}
\right)
\left(
h_{i}-\bar{h}
\right)
}{
\sqrt{
\sum_{i=1}^{N}
\left(
\hat{h}_{i}-\bar{\hat{h}}
\right)^{2}
}
\sqrt{
\sum_{i=1}^{N}
\left(
h_{i}-\bar{h}
\right)^{2}
}
}.
\label{eq:hr_correlation}
\end{equation}
Where $\bar{\hat{h}}$ and $\bar{h}$ denote the mean estimated and
ground-truth heart rates, respectively. Lower MAE and RMSE indicate
smaller estimation errors, while a higher $\rho$ indicates stronger
agreement between the estimated and ground-truth heart rates.

\begin{table*}[t]
  \caption{Intra-dataset comparison of remote HR estimation on NIRP-IND,
  NIRP-DRV, VIPL-HR, BUAA-MIHR, and MMPD. MAE and RMSE are reported
  in bpm. $\downarrow$ indicates that lower values are better, while
  $\uparrow$ indicates that higher values are better. The best results
  are highlighted in \textbf{bold} with a light-lavender background,
  while the second-best results are marked with a light-apricot
  background.}
  \label{tab:intra_dataset}

  \centering
  \scriptsize
  \renewcommand{\arraystretch}{1.08}
  \setlength{\tabcolsep}{0.65pt}

  \begin{tabularx}{\textwidth}
  {@{}>{\raggedright\arraybackslash}p{0.205\textwidth}*{15}{Y}@{}}
    \toprule

    \multirow{2}{*}{\makecell[l]{Method / Venue}} &
    \multicolumn{3}{c}{NIRP-IND} &
    \multicolumn{3}{c}{NIRP-DRV} &
    \multicolumn{3}{c}{VIPL-HR} &
    \multicolumn{3}{c}{BUAA-MIHR} &
    \multicolumn{3}{c}{MMPD} \\

    \cmidrule(lr){2-4}
    \cmidrule(lr){5-7}
    \cmidrule(lr){8-10}
    \cmidrule(lr){11-13}
    \cmidrule(lr){14-16}

    &
    MAE$\downarrow$ & RMSE$\downarrow$ & $\rho\uparrow$ &
    MAE$\downarrow$ & RMSE$\downarrow$ & $\rho\uparrow$ &
    MAE$\downarrow$ & RMSE$\downarrow$ & $\rho\uparrow$ &
    MAE$\downarrow$ & RMSE$\downarrow$ & $\rho\uparrow$ &
    MAE$\downarrow$ & RMSE$\downarrow$ & $\rho\uparrow$ \\

    \midrule

    POS \cite{63} / \textit{TBME'2016}
    & 5.52 & 6.85 & 0.40
    & 12.75 & 15.36 & 0.34
    & 11.50 & 17.20 & 0.30
    & 5.04 & 7.12 & 0.63
    & 12.36 & 17.71 & 0.18 \\

    CHROM \cite{chrom} / \textit{TBME'2013}
    & 6.84 & 8.41 & 0.32
    & 14.52 & 17.41 & 0.18
    & 11.07 & 17.99 & 0.27
    & 6.09 & 8.29 & 0.51
    & 13.66 & 18.76 & 0.08 \\

    LGI \cite{LGI} / \textit{CVPR'2018}
    & 8.95 & 11.01 & 0.39
    & 12.90 & 15.51 & 0.31
    & 12.84 & 19.02 & 0.29
    & 6.97 & 11.33 & 0.42
    & 17.08 & 23.32 & 0.04 \\

    DeepPhys \cite{DeepPhys} / \textit{ECCV'2018}
    & 3.11 & 4.44 & 0.74
    & 13.22 & 18.39 & 0.43
    & 11.00 & 13.80 & 0.11
    & 4.78 & 6.74 & 0.69
    & 22.27 & 28.92 & -0.03 \\

    TS-CAN \cite{TS-CAN} / \textit{NeurIPS'2020}
    & 2.49 & 3.89 & 0.75
    & 12.70 & 18.03 & 0.47
    & 9.39 & 14.59 & 0.21
    & 4.84 & 6.89 & 0.68
    & 9.71 & 17.22 & 0.44 \\

    PFE-TFA \cite{PFE-TFA} / \textit{AAAI'2023}
    & 2.81 & 4.57 & 0.84
    & 5.34 & 8.92 & 0.73
    & 6.91 & 8.65 & 0.65
    & 1.29 & 2.65 & 0.91
    & -- & -- & -- \\

    NEST \cite{NEST} / \textit{CVPR'2023}
    & 1.08 & 2.26 & 0.89
    & 3.61 & 7.32 & 0.82
    & 5.52 & 7.96 & 0.80
    & 2.88 & 4.69 & 0.89
    & -- & -- & -- \\

    ND-DeeprPPG \cite{13} / \textit{TIP'2024}
    & 0.66 & 1.45 & 0.89
    & 3.47 & 6.54 & 0.85
    & 5.15 & 7.52 & 0.78
    & \cellcolor{secondbest}0.58
    & 1.81
    & \cellcolor{secondbest}0.95
    & -- & -- & -- \\

    EfficientPhys \cite{Efficientphys} / \textit{WACV'2023}
    & 1.37 & 4.81 & 0.81
    & 3.67 & 12.28 & 0.81
    & 5.23 & 8.25 & 0.72
    & 1.43 & 4.98 & 0.93
    & 13.47 & 21.32 & 0.21 \\

    PhysFormer++ \cite{Physformer++} / \textit{IJCV'2023}
    & 0.64 & 1.39 & 0.89
    & 3.56 & 7.59 & 0.83
    & 4.88 & 7.62 & 0.80
    & 0.93 & 1.66 & 0.91
    & -- & -- & -- \\

    RhythmFormer \cite{Rhythmformer} / \textit{PR'2025}
    & 0.52
    & 1.10
    & \cellcolor{secondbest}0.91
    & 3.44 & 6.72 & 0.82
    & 4.75 & 7.49 & 0.82
    & 0.67 & 1.57 & 0.94
    & 4.69 & 11.31 & 0.60 \\

    PhysDiff \cite{PhysDiff} / \textit{AAAI'2025}
    & 0.50
    & 1.06
    & \cellcolor{secondbest}0.91
    & 3.24 & 6.26 & 0.83
    & 3.92 & 6.65 & 0.85
    & 0.82 & 1.71 & 0.93
    & 7.17 & 9.63 & 0.78 \\

    PHASE-Net \cite{PHASE-Net} / \textit{CVPR'2026}
    & \cellcolor{secondbest}0.48
    & \cellcolor{secondbest}0.98
    & \cellcolor{secondbest}0.91
    & 3.18
    & 6.23
    & 0.86
    & \cellcolor{secondbest}3.87
    & \cellcolor{secondbest}6.37
    & \cellcolor{secondbest}0.86
    & 0.63
    & \cellcolor{secondbest}1.43
    & 0.94
    & 4.78 & 8.22 & 0.71 \\

    PhysLLM \cite{Physllm} / \textit{ICLR'2026}
    & 0.57
    & 1.27
    & 0.90
    & \cellcolor{secondbest}2.95
    & \cellcolor{secondbest}6.13
    & \cellcolor{secondbest}0.86
    & 4.24
    & 6.81
    & 0.85
    & 0.78
    & 1.62
    & 0.94
    & \cellcolor{secondbest}4.36 & \cellcolor{secondbest}10.76 & \cellcolor{secondbest}0.65 \\

    \rowcolor{bestrow}
    \textbf{PhysVR (Ours)}
    & \textbf{0.38}
    & \textbf{0.85}
    & \textbf{0.92}
    & \textbf{2.56}
    & \textbf{5.85}
    & \textbf{0.87}
    & \textbf{3.59}
    & \textbf{5.68}
    & \textbf{0.88}
    & \textbf{0.42}
    & \textbf{1.09}
    & \textbf{0.96}
    & \textbf{3.92}
    & \textbf{8.03}
    & \textbf{0.80} \\

    \bottomrule
  \end{tabularx}
\end{table*}

\subsection{Implementation Details}

We implement PhysVR using the rPPG-Toolbox~\cite{rppg-toolbox} and
conduct all experiments on RGB facial videos. Facial regions are cropped,
resized to $128 \times 128$, and divided into 160-frame clips after
temporal alignment and reference-signal resampling. For each clip, eight
uniformly sampled frames are processed with a fixed interference-oriented
prompt by the frozen Qwen2-VL-2B-Instruct~\cite{qwen2}. All remaining
components are jointly optimized for 30 epochs using AdamW with a learning
rate of $3 \times 10^{-4}$. The shared correction scale
$\lambda_{\mathrm{shr}}$ is initialized to $0.10$, while the four
expert-specific scales $\lambda_{\mathrm{ill}}$,
$\lambda_{\mathrm{mot}}$, $\lambda_{\mathrm{blur}}$, and
$\lambda_{\mathrm{roi}}$ are initialized to $0.05$. We set
$\lambda_{\mathrm{mse}}=1.00$, $\lambda_{\mathrm{np}}=0.25$,
$\lambda_{\mathrm{freq}}=0.05$, and $\lambda_{\mathrm{c}}=0.30$.

All dataset partitions are subject-disjoint. VIPL-HR follows the official
five-fold protocol, with four folds used for training and one for testing.
BUAA-MIHR, MMPD, and the combined NIRP dataset use a $3{:}1$
training-to-testing split. NIRP-IND and NIRP-DRV are partitioned jointly
but evaluated separately, while BUAA-MIHR and MMPD are used only for
intra-dataset evaluation. Following prior work~\cite{1}, cross-dataset
evaluation is performed without target-domain training or fine-tuning.
We report results for NIRP $\rightarrow$ VIPL-HR,
VIPL-HR $\rightarrow$ NIRP-IND, and
VIPL-HR $\rightarrow$ NIRP-DRV.

\begin{table*}[t]
  \caption{Cross-dataset comparison of remote HR estimation. The models are
  trained on a source dataset and evaluated on a different target dataset.
  MAE and RMSE are reported in bpm. $\downarrow$ indicates that lower values
  are better, while $\uparrow$ indicates that higher values are better.
  The best results are highlighted in \textbf{bold} with a light-lavender
  background, while the second-best results are marked with a light-apricot
  background.}
  \label{tab:cross_dataset}

  \centering
  \footnotesize
  \renewcommand{\arraystretch}{1.08}
  \setlength{\tabcolsep}{2.2pt}

  \begin{tabularx}{\textwidth}
  {@{}>{\raggedright\arraybackslash}p{0.25\textwidth}*{9}{Y}@{}}
    \toprule

    \multirow{2}{*}{\makecell[l]{Method / Venue}} &
    \multicolumn{3}{c}{NIRP $\rightarrow$ VIPL-HR} &
    \multicolumn{3}{c}{VIPL-HR $\rightarrow$ NIRP-IND} &
    \multicolumn{3}{c}{VIPL-HR $\rightarrow$ NIRP-DRV} \\

    \cmidrule(lr){2-4}
    \cmidrule(lr){5-7}
    \cmidrule(lr){8-10}

    &
    MAE$\downarrow$ & RMSE$\downarrow$ & $\rho\uparrow$ &
    MAE$\downarrow$ & RMSE$\downarrow$ & $\rho\uparrow$ &
    MAE$\downarrow$ & RMSE$\downarrow$ & $\rho\uparrow$ \\

    \midrule

    DeepPhys \cite{DeepPhys} / \textit{ECCV'2018}
    & 15.53 & 17.48 & 0.41
    & 6.58 & 9.16 & 0.52
    & 10.51 & 12.71 & 0.56 \\

    TS-CAN \cite{TS-CAN} / \textit{NeurIPS'2020}
    & 12.04 & 15.12 & 0.45
    & 6.57 & 9.25 & 0.57
    & 10.37 & 12.65 & 0.61 \\

    PFE-TFA \cite{PFE-TFA} / \textit{AAAI'2023}
    & 8.20 & 11.22 & 0.64
    & 1.87 & 3.67 & 0.77
    & 7.82 & 9.99 & 0.64 \\

    NEST \cite{NEST} / \textit{CVPR'2023}
    & 5.15 & 8.78 & 0.72
    & 2.78 & 4.63 & 0.78
    & 6.27 & 8.45 & 0.65 \\

    ND-DeeprPPG \cite{13} / \textit{TIP'2024}
    & 5.08
    & \cellcolor{secondbest}7.92
    & \cellcolor{secondbest}0.75
    & 1.74
    & 3.28
    & 0.81
    & 6.18
    & 7.75
    & 0.68 \\

    EfficientPhys \cite{Efficientphys} / \textit{WACV'2023}
    & 5.25 & 11.81 & 0.71
    & 1.80 & 6.19 & 0.76
    & 6.07 & 11.53 & 0.64 \\

    PhysFormer++ \cite{Physformer++} / \textit{IJCV'2023}
    & 5.09 & 8.97 & 0.71
    & 1.79 & 3.77 & 0.81
    & 5.78 & 8.65 & 0.66 \\

    RhythmFormer \cite{Rhythmformer} / \textit{PR'2025}
    & 4.83 & 8.05 & 0.74
    & 1.70 & 3.43 & 0.83
    & 5.40 & 7.93 & 0.71 \\

    PhysDiff \cite{PhysDiff} / \textit{AAAI'2025}
    & 4.34
    & 8.12
    & 0.74
    & 1.62
    & 3.35
    & 0.85
    & 5.46
    & 7.96
    & \cellcolor{secondbest}0.72 \\

    PHASE-Net \cite{PHASE-Net} / \textit{CVPR'2026}
    & \cellcolor{secondbest}4.27
    & 7.96
    & 0.75
    & \cellcolor{secondbest}1.53
    & \cellcolor{secondbest}3.22
    & \cellcolor{secondbest}0.86
    & \cellcolor{secondbest}5.39
    & \cellcolor{secondbest}7.71
    & \cellcolor{secondbest}0.72 \\

    PhysLLM \cite{Physllm} / \textit{ICLR'2026}
    & 4.51
    & 8.37
    & 0.73
    & 1.69
    & 3.49
    & 0.85
    & 5.56
    & 8.15
    & 0.71 \\

    \rowcolor{bestrow}
    \textbf{PhysVR (Ours)}
    & \textbf{3.92}
    & \textbf{7.23}
    & \textbf{0.77}
    & \textbf{1.34}
    & \textbf{3.07}
    & \textbf{0.87}
    & \textbf{5.14}
    & \textbf{7.21}
    & \textbf{0.74} \\

    \bottomrule
  \end{tabularx}
\end{table*}

\begin{figure}[t]
  \centering
  \setlength{\fboxsep}{0pt}
  \includegraphics[width=\linewidth]{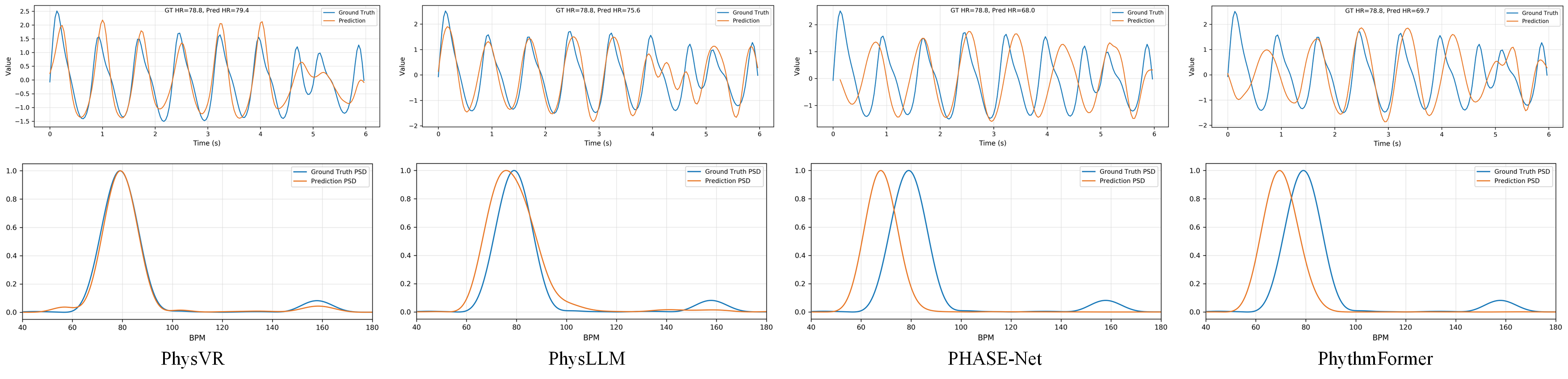}
  \caption{Comparison of rPPG estimation results on VIPL-HR. The top row presents the predicted and ground-truth rPPG signals, while the bottom row shows their corresponding normalized power spectral density distributions. From left to right, the columns correspond to PhysVR, PhysLLM, PHASE-Net, and RhythmFormer.}
  \label{fig:33}
\end{figure}

\subsection{Intra-dataset Evaluation}
\label{sec:intra_dataset}

Table~\ref{tab:intra_dataset} summarizes the intra-dataset results on
five benchmarks. PhysVR achieves the best overall performance across all
evaluated datasets, consistently yielding lower MAE and RMSE together
with higher correlation. On NIRP-IND, PhysVR obtains an MAE of
0.38~bpm, an RMSE of 0.85~bpm, and a correlation coefficient of 0.92.
Compared with PHASE-Net, the strongest competing method on this subset,
the MAE and RMSE are reduced by 20.8\% and 13.3\%, respectively.
On the more challenging NIRP-DRV subset, PhysVR achieves an MAE of
2.56~bpm and an RMSE of 5.85~bpm, compared with 2.95~bpm and
6.13~bpm for PhysLLM, while improving the correlation coefficient from
0.86 to 0.87. These results demonstrate robust estimation performance
under both indoor and driving conditions. PhysVR also achieves the best results on VIPL-HR, BUAA-MIHR, and MMPD. On VIPL-HR, it obtains an MAE of 3.59~bpm, an RMSE of
5.68~bpm, and a correlation coefficient of 0.88, reducing the MAE and
RMSE of PHASE-Net by 7.2\% and 10.8\%, respectively. On BUAA-MIHR,
the best competing MAE and RMSE are reduced from 0.58~bpm and
1.43~bpm to 0.42~bpm and 1.09~bpm, corresponding to improvements
of 27.6\% and 23.8\%, respectively, while the correlation coefficient
reaches 0.96. On MMPD, PhysVR obtains an MAE of 3.92~bpm, an RMSE
of 8.03~bpm, and a correlation coefficient of 0.80. Compared with the
best competing results, the MAE and RMSE are reduced by 10.1\% and
2.3\%, respectively. Overall, the consistent improvements across these
benchmarks demonstrate the robustness of PhysVR under diverse conditions.

\begin{figure}[t]
  \centering
  \setlength{\fboxsep}{0pt}
  \includegraphics[width=\linewidth]{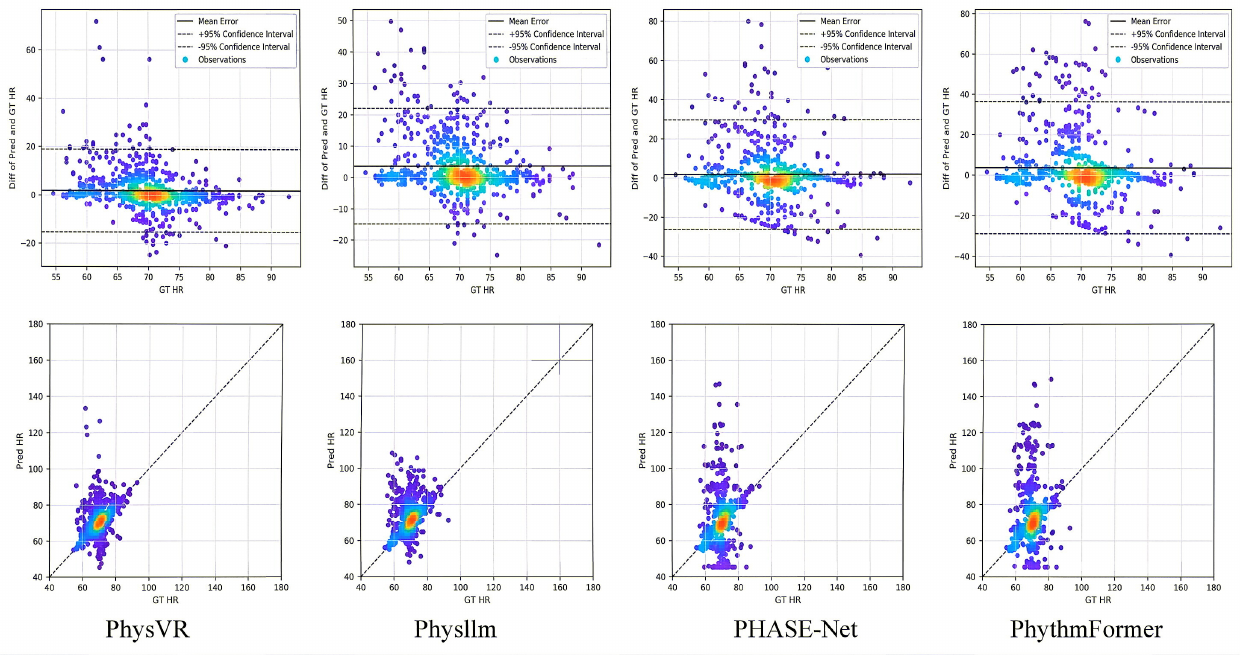}
  \caption{Comparison of HR estimation results on NIRP-DRV. The top row presents the difference between the predicted and ground-truth HR values, while the bottom row shows the corresponding scatter plots of predicted HR against ground-truth HR. The diagonal line represents perfect agreement. From left to right, the columns correspond to PhysVR, PhysLLM, PHASE-Net, and RhythmFormer.}
  \label{fig:34}
\end{figure}

Beyond the aggregate metrics, Fig.~\ref{fig:33} compares the predicted
rPPG signals and power spectra on VIPL-HR. For the representative sample,
PhysVR estimates an HR of 79.4~bpm, close to the ground-truth value of
78.8~bpm, with an absolute error of 0.6~bpm. Its predicted signal follows
the reference periodicity, while the dominant spectral peak is well aligned
with the ground-truth HR frequency. In contrast, PhysLLM, PHASE-Net,
and RhythmFormer exhibit larger temporal deviations and spectral shifts.
Fig.~\ref{fig:34} further compares HR estimates on NIRP-DRV,
where PhysVR shows a tighter error distribution around zero, with
predictions more concentrated around the identity line. These observations
are consistent with the quantitative results in Table~\ref{tab:intra_dataset}
and further support the effectiveness of PhysVR under challenging
recording conditions.

\begin{figure}[t]
  \centering
  \setlength{\fboxsep}{0pt}
  \includegraphics[width=\linewidth]{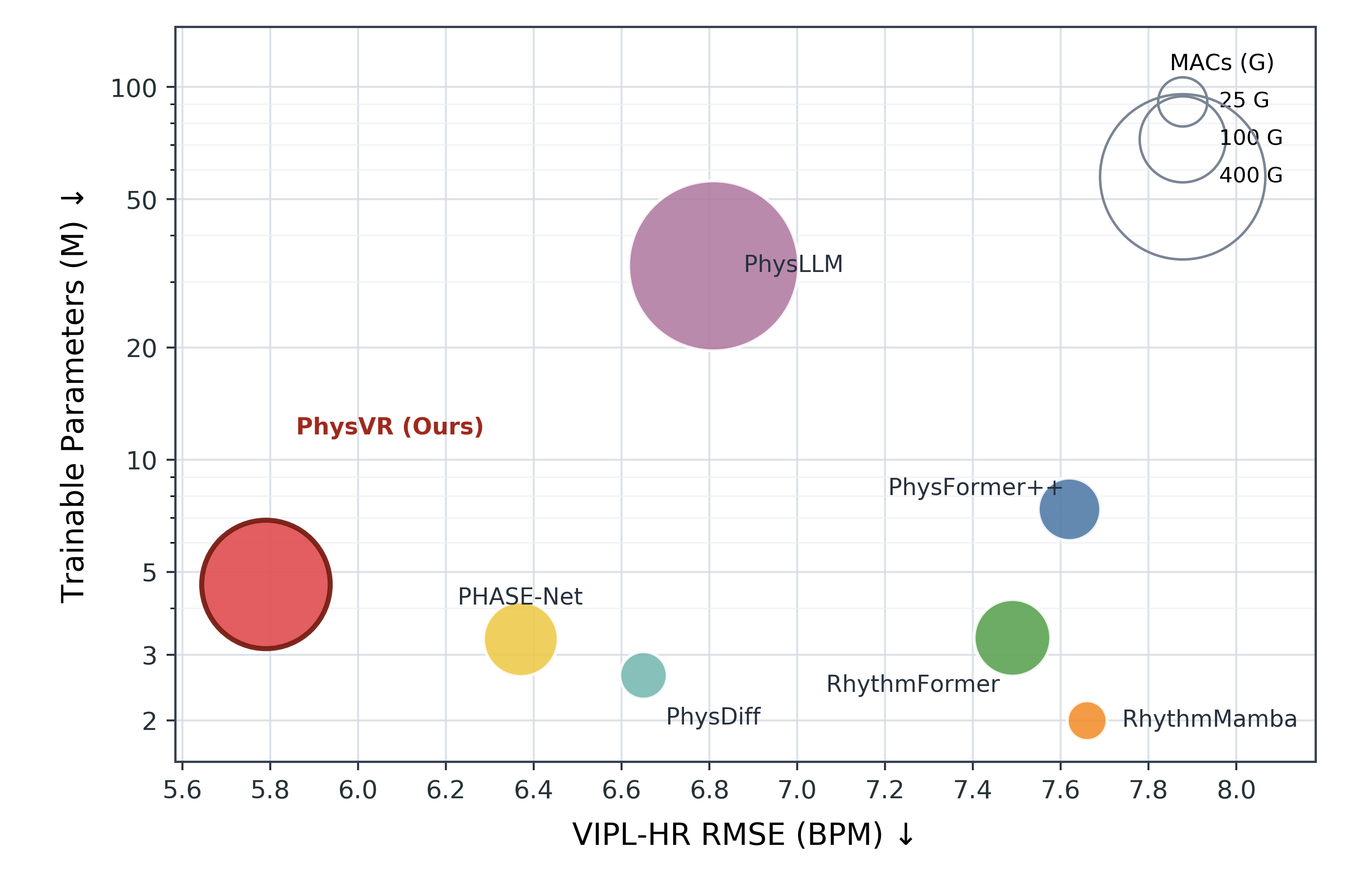}
\caption{Comparison of HR estimation accuracy and computational
complexity on VIPL-HR. The horizontal axis denotes RMSE, the vertical
axis denotes the number of trainable parameters, and the bubble size
represents MACs. For PhysVR, the MACs include the forward computation
of the frozen VLM on eight uniformly sampled frames from each clip.}
  \label{fig:37}
\end{figure}

\subsection{Cross-dataset Evaluation}
\label{sec:cross_dataset}

Table~\ref{tab:cross_dataset} summarizes the cross-dataset results
between NIRP and VIPL-HR. PhysVR achieves the best overall performance
across all three transfer settings. When trained on NIRP and evaluated
on VIPL-HR, PhysVR obtains an MAE of 3.92~bpm, an RMSE of 7.23~bpm,
and a correlation coefficient of 0.77. Compared with the best competing
value for each metric, the MAE and RMSE are reduced by 8.2\% and 8.7\%,
respectively, while the correlation coefficient increases from 0.75
to 0.77. When trained on VIPL-HR and evaluated on NIRP-IND, PhysVR achieves
an MAE of 1.34~bpm, an RMSE of 3.07~bpm, and a correlation coefficient
of 0.87. Compared with PHASE-Net, the MAE and RMSE are reduced by
12.4\% and 4.7\%, respectively. The improvement remains evident on
challenging NIRP-DRV dataset, where PhysVR achieves an MAE
of 5.14~bpm, an RMSE of 7.21~bpm, and a correlation coefficient of
0.74, compared with 5.39~bpm, 7.71~bpm, and 0.72 for PHASE-Net.
These results show that PhysVR maintains high HR estimation performance across datasets.

Together with the intra-dataset results, the cross-dataset evaluation
further demonstrates the robustness of PhysVR across different recording
conditions and data distributions. The consistent generalization
performance also supports the effectiveness of interference-aware
temporal feature refinement. Fig.~\ref{fig:37} further compares HR estimation accuracy and
computational complexity on VIPL-HR. PhysVR achieves the lowest RMSE
among the compared methods. Although PhysVR employs a frozen
2B-parameter VLM, only eight uniformly sampled frames from each clip
are processed by the VLM, and its forward computation is included in
the reported MACs. This restricted VLM processing limits the additional
computational overhead while retaining the performance gains brought
by interference-aware temporal feature refinement.

\begin{figure}[t]
  \centering
  \setlength{\fboxsep}{0pt}
  \includegraphics[width=\linewidth]{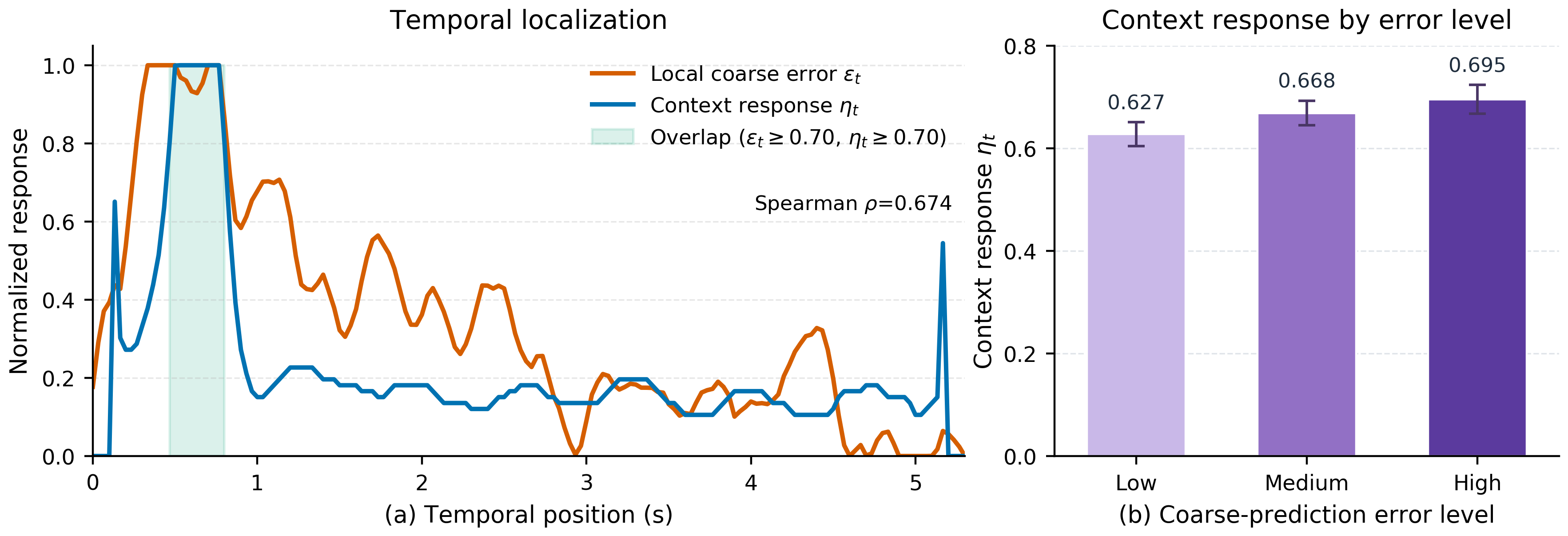}
\caption{Temporal response of the interference-aware context on
BUAA-MIHR. (a) Temporal correspondence between local coarse-prediction
error and context response, with shaded high-response regions.
(b) Average context response for clips with low, medium, and high
coarse-prediction errors. Error bars indicate 95\% confidence intervals.}
  \label{fig:41}
\end{figure}

\subsection{Ablation Studies} 

\textbf{Temporal Response of the Interference-Aware Context:}
We investigate whether the interference-aware temporal context exhibits
stronger responses at temporal positions with larger coarse-prediction
errors. The local prediction error is computed from the normalized
deviation between the coarse prediction and the ground-truth signal,
while the context response is quantified by the normalized magnitude of
$\mathbf{C}$ at each temporal position. As shown in
Fig.~\ref{fig:41}(a), the two sequences exhibit clear temporal
correspondence, with a Spearman correlation of $0.674$. We further
divide the test clips into three equal-sized groups according to their
coarse-prediction RMSE. Fig.~\ref{fig:41}(b) shows that the average
context response increases from $0.627$ in the low-error group to
$0.695$ in the high-error group. These results indicate that the
constructed temporal context responds more strongly when the coarse
prediction is less reliable, supporting its role in guiding subsequent
feature refinement.

\begin{figure}[t]
  \centering
  \setlength{\fboxsep}{0pt}
  \includegraphics[width=\linewidth]{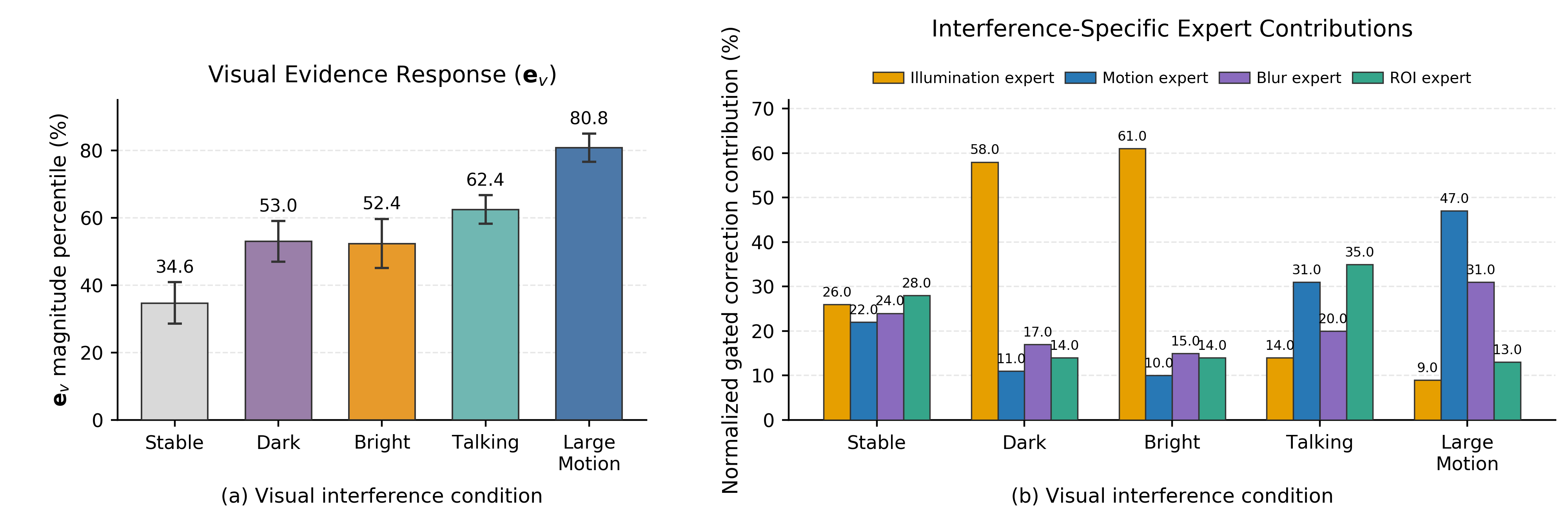}
\caption{Analysis of visual evidence and interference-specific expert
contributions on VIPL-HR. (a) Direct prediction estimates rPPG from learned temporal features without further refinement.
(b) PhysVR combines physiological reliability evidence with VLM-derived visual interference evidence to guide temporal feature refinement before final estimation.
(c) An example showing that learned temporal features may remain affected by interference under challenging visual conditions, while further refinement improves the final rPPG prediction.}
  \label{fig:40}
\end{figure}

\textbf{Analysis of Visual Evidence and Interference-Specific Experts:}
Fig.~\ref{fig:40} analyzes the responses of the visual interference
evidence and the four interference-specific experts on VIPL-HR.
As shown in Fig.~\ref{fig:40}(a), the percentile rank of the
visual-evidence magnitude is lowest under the stable condition at
$34.6\%$ and increases under interference conditions, reaching
$80.8\%$ under large motion. This result indicates that the
VLM-derived visual evidence responds more strongly when visual
interference is present. Fig.~\ref{fig:40}(b) further compares the
normalized gated correction contributions of the four experts.
The illumination expert shows the largest contribution under dark
and bright conditions, accounting for $58.0\%$ and $61.0\%$,
respectively, while the motion expert contributes most strongly
under large motion at $47.0\%$. Under talking, the contributions
are more distributed, with the ROI-instability, motion, and blur
experts contributing $35.0\%$, $31.0\%$, and $20.0\%$.
These results demonstrate condition-dependent responses of the
interference-specific experts to different interference.

\begin{figure}[t]
  \centering
  \setlength{\fboxsep}{0pt}
  \includegraphics[width=\linewidth]{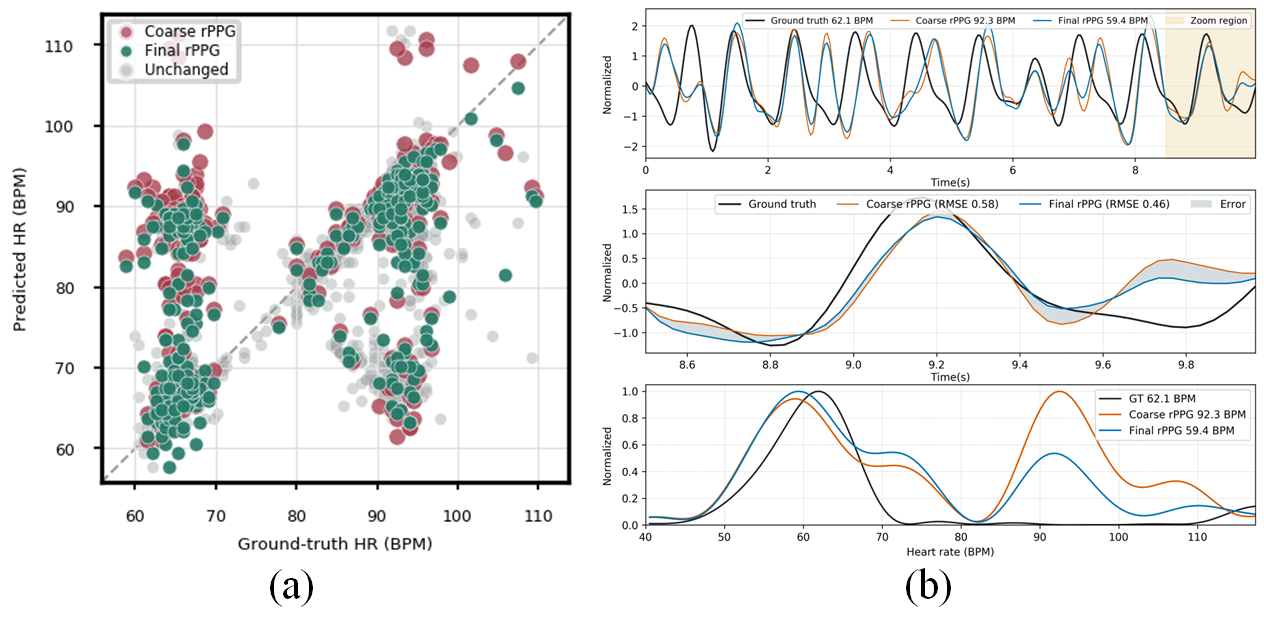}
  \caption{Effect of temporal feature refinement on BUAA-MIHR.
  (a) Comparison of coarse and final HR estimates. Gray markers denote
  unchanged estimates. (b) Temporal and spectral comparison of the
  ground-truth, coarse, and final rPPG signals for a representative
  sample. The panels show the complete sequence, an enlarged interval,
  and the normalized power spectra from top to bottom.}
  \label{fig:38}
\end{figure}

\textbf{Effectiveness of Temporal Feature Refinement:}
To examine whether learned temporal features benefit from further refinement
after backbone feature learning, Fig.~\ref{fig:38} compares the coarse and
final predictions of PhysVR on BUAA-MIHR.
As shown in Fig.~\ref{fig:38}(a), the final HR estimates are
closer to the identity line than the coarse estimates.
Fig.~\ref{fig:38}(b) presents a representative example in the
temporal and frequency domains. Within the enlarged interval, the local
RMSE decreases from $0.58$ to $0.46$, while the dominant spectral peak
shifts from $92.3$ to $59.4$~bpm, approaching the ground-truth HR of
$62.1$~bpm. These results show that further temporal feature refinement
after backbone feature learning improves rPPG estimation.

\textbf{Ablation of Physiological and Visual Evidence:}
Table~\ref{tab:evidence_ablation} evaluates the contributions of the
physiological reliability evidence $\mathbf{R}$ and visual interference
evidence $\mathbf{e}_{v}$ to temporal feature refinement. The refinement
architecture is kept unchanged, with only the corresponding evidence
input disabled. The coarse prediction obtains an MAE of $3.79$~bpm
and an RMSE of $6.83$~bpm. Applying temporal feature refinement without either evidence source slightly reduces the MAE and RMSE from $3.79$ and $6.83$~bpm to $3.67$ and $6.72$~bpm, respectively, indicating that refinement alone provides a modest improvement over the coarse prediction. Introducing physiological evidence reduces
the MAE and RMSE to $3.26$ and $6.31$~bpm, while visual evidence achieves
$3.15$ and $6.27$~bpm, respectively. Combining both evidence sources
yields the best performance, with an MAE of $2.56$~bpm and an RMSE of
$5.85$~bpm. These results show that both evidence sources contribute to
temporal feature refinement, with their combination providing the largest
performance improvement.

\textbf{Ablation of the Context-Guided Refinement Module:}
Table~\ref{tab:refinement_ablation} evaluates the contribution of each
component in the context-guided refinement module. Removing the entire
refinement module increases the MAE and RMSE from $3.59/5.68$~bpm to
$4.82/6.87$~bpm, demonstrating the necessity of refining learned temporal
features before final estimation. Among the component-level variants,
removing the shared correction unit causes the largest performance
degradation, highlighting the importance of general temporal correction
before interference-specific refinement. Removing any of the four
interference-specific experts also degrades performance, with MAE ranging
from $4.29$ to $4.41$~bpm and RMSE from $6.27$ to $6.39$~bpm, indicating
that each expert provides complementary refinement capability. Finally,
disabling adaptive routing increases the MAE and RMSE to $4.16$ and
$6.08$~bpm, respectively, demonstrating the benefit of adaptive expert
selection. Overall, the full model achieves the best performance,
supporting the effectiveness of the complete context-guided refinement
design.

\textbf{Interference-Specific Expert Specialization:}
Table~\ref{tab:condition_expert_ablation} further investigates whether
the four experts exhibit specialization toward different types of
interference. The illumination expert shows the strongest contribution
under dark and bright conditions, as removing it results in the largest
performance degradation in both cases. Under large motion, the motion
expert becomes the most influential, while removing the blur expert also
causes a clear performance drop. For talking sequences, the
ROI-instability expert plays the most prominent role. These results show
that the four experts contribute differently under different interference
conditions, supporting their interference-specific design.

\begin{table}[t]
    \centering
\caption{Ablation of physiological and visual evidence for temporal feature refinement on NIRP-DRV.}
    \label{tab:evidence_ablation}
    \small
    \setlength{\tabcolsep}{2.2pt}
    \renewcommand{\arraystretch}{1.12}
    \begin{tabular*}{\columnwidth}
        {@{\extracolsep{\fill}}lccccc@{}}
        \toprule
        Configuration
        & Refinement
        & $\mathbf{R}$
        & $\mathbf{e}_{v}$
        & MAE $\downarrow$
        & RMSE $\downarrow$ \\
        \midrule
        Coarse prediction
        & \xmark
        & \xmark
        & \xmark
        & 3.79
        & 6.83 \\

                Feature-only refinement
                & \cmark
                & \xmark
                & \xmark
                & 3.67
                & 6.72 \\

        $+$ Physiological evidence
        & \cmark
        & \cmark
        & \xmark
        & 3.26
        & 6.31 \\

        $+$ Visual evidence
        & \cmark
        & \xmark
        & \cmark
        & 3.15
        & 6.27 \\

        \midrule
        \textbf{Full model}
        & \cmark
        & \cmark
        & \cmark
        & \textbf{2.56}
        & \textbf{5.85} \\
        \bottomrule
    \end{tabular*}
\end{table}

\begin{table}[t]
\centering
\caption{Ablation of the context-guided refinement module in PhysVR on VIPL-HR.}
\label{tab:refinement_ablation}
\small
\renewcommand{\arraystretch}{1.12}
\begin{tabular*}{\columnwidth}
{@{\extracolsep{\fill}}lcc@{}}
\toprule
Configuration
& MAE $\downarrow$
& RMSE $\downarrow$ \\
\midrule

No feature refinement
& 4.82
& 6.87 \\

w/o Shared correction unit
& 4.75
& 6.78 \\

w/o Illumination expert
& 4.37
& 6.35 \\

w/o Motion expert
& 4.29
& 6.27 \\

w/o Blur expert
& 4.35
& 6.32 \\

w/o ROI-instability expert
& 4.41
& 6.39 \\

w/o Adaptive routing
& 4.16
& 6.08 \\

\midrule
\textbf{Full model}
& \textbf{3.59}
& \textbf{5.68} \\

\bottomrule
\end{tabular*}
\end{table}

\begin{table}[t]
    \centering
    \caption{Condition-wise ablation of interference-specific experts on
    VIPL-HR.}
    \label{tab:condition_expert_ablation}
    \small
    \setlength{\tabcolsep}{3.0pt}
    \renewcommand{\arraystretch}{1.12}

    \begin{tabular*}{\columnwidth}
        {@{\extracolsep{\fill}}lcccc@{}}
        \toprule
        Removed expert
        & Dark
        & Bright
        & Talking
        & Motion \\
        \midrule

        Illumination
        & 5.02/7.22
        & 4.88/7.05
        & 4.28/6.42
        & 4.76/7.08 \\

        Motion
        & 4.15/6.34
        & 3.82/6.03
        & 4.72/6.86
        & 5.63/8.04 \\

        Blur
        & 4.17/6.37
        & 3.85/6.06
        & 4.49/6.68
        & 5.24/7.64 \\

        ROI-instability
        & 4.20/6.39
        & 3.88/6.10
        & 5.08/7.28
        & 4.97/7.26 \\

        \midrule

        \textbf{Full model}
        & \textbf{3.92/6.05}
        & \textbf{3.61/5.77}
        & \textbf{4.08/6.16}
        & \textbf{4.54/6.83} \\

        \bottomrule
    \end{tabular*}
\end{table}

\begin{table}[t]
    \centering
    \caption{Comparison of visual encoders, prompt designs, and frozen VLM
    backbones on the VIPL-HR dataset.}
    \label{tab:vlm_prompt_ablation}
    \small
    \setlength{\tabcolsep}{3.2pt}
    \renewcommand{\arraystretch}{1.12}
    \begin{tabular*}{\columnwidth}
        {@{\extracolsep{\fill}}llcc@{}}
        \toprule
        Encoder
        & Prompt 
        & MAE $\downarrow$
        & RMSE $\downarrow$ \\
        \midrule

        DINOv3~\cite{dinov3}
        & \xmark
        & 4.49
        & 6.65 \\

        PE~\cite{perception}
        & \xmark
        & 4.27
        & 6.48 \\

        \midrule

        Qwen2-VL-2B~\cite{qwen2}
        & Generic
        & 3.83
        & 5.92 \\

        Qwen2-VL-2B~\cite{qwen2}
        & Interference-oriented
        & 3.59
        & 5.68 \\

        \midrule

        InternVL2.5-2B~\cite{interbvl25}
        & Interference-oriented
        & 3.69
        & 5.85 \\

        SmolVLM2-2.2B~\cite{smolvlm}
        & Interference-oriented
        & 3.51
        & 5.63 \\

        Qwen2-VL-7B~\cite{qwen2}
        & Interference-oriented
        & 3.47
        & 5.61 \\

        \bottomrule
    \end{tabular*}
\end{table}

\textbf{Effect of Visual Encoder, Prompt Design, and VLM Backbone:}
Table~\ref{tab:vlm_prompt_ablation} evaluates the effects of visual
encoders, prompt designs, and VLM backbones on VIPL-HR. Compared with
frozen visual encoders such as DINOv3 and PE, the VLM-based evidence
branch achieves better performance, suggesting that language-guided
visual representations provide more effective interference-related
guidance for temporal feature refinement. With Qwen2-VL-2B fixed,
replacing the generic prompt with the interference-oriented prompt
reduces the MAE from $3.83$ to $3.59$~bpm and the RMSE from $5.92$ to
$5.68$~bpm, showing that explicitly directing the VLM toward
interference-related factors provides more useful visual evidence.
We further compare Qwen2-VL-2B with InternVL2.5-2B, SmolVLM2-2.2B,
and Qwen2-VL-7B using the same prompt. Qwen2-VL-2B outperforms
InternVL2.5-2B, while SmolVLM2-2.2B further reduces the MAE and RMSE
to $3.51$ and $5.63$~bpm, respectively. Qwen2-VL-7B achieves the best
performance with an MAE of $3.47$~bpm and an RMSE of $5.61$~bpm, while
the gain over smaller backbones remains marginal despite the increased
model size.
\section{Conclusion}
\label{sec:conclusion}

In this work, we presented PhysVR, a vision-language model guided
interference-aware temporal feature refinement framework for robust rPPG
estimation. PhysVR combines time-resolved physiological reliability evidence
with clip-level visual interference evidence to guide the refinement of learned
temporal features, enabling further interference suppression before final rPPG
estimation. Extensive experiments on five public benchmarks demonstrate
consistent improvements under both intra-dataset and cross-dataset settings,
while ablation studies further verify the effectiveness of the proposed
refinement design. Although the frozen VLM
introduces additional computational overhead, future work will investigate
knowledge distillation and more compact visual interference perception models
to further improve efficiency.

\section*{Acknowledgments}
This work was supported by the key project of Jiangsu Provincial Natural Science Fund under Grant No. BK20253028 and the National Nature Science Fund of China under Grant Nos. 62176124, U24A20330 and 62361166670.

\bibliographystyle{IEEEtran}
\bibliography{refernce}

@String(CVPR= {IEEE Conf. Comput. Vis. Pattern Recog.})

@String(ECCV= {Eur. Conf. Comput. Vis.})

@String(AAAI = {AAAI})

@String(CVPR  = {CVPR})

@String(ECCV  = {ECCV})

@String{Computing = "Computing" }

@String{Computer = "{IEEE} Computer" }

@String{Springer = "Springer-Verlag" }

@inproceedings{1,
  title={Remote Photoplethysmography in Real-World and Extreme Lighting Scenarios},
  author={Shao, Hang and Luo, Lei and Qian, Jianjun and Yan, Mengkai and Chen, Shuo and Yang, Jian},
  booktitle={Proceedings of the Computer Vision and Pattern Recognition Conference},
  pages={10858--10867},
  year={2025}
}

@inproceedings{2,
  title={Continual Learning for Remote Physiological Measurement: Minimize Forgetting and Simplify Inference},
  author={Liang, Qian and Chen, Yan and Hu, Yang},
  booktitle={European conference on computer vision},
  pages={126--144},
  year={2024},
  organization={Springer}
}

@article{3,
  title={A novel algorithm for remote photoplethysmography: Spatial subspace rotation},
  author={Wang, Wenjin and Stuijk, Sander and De Haan, Gerard},
  journal={IEEE transactions on biomedical engineering},
  volume={63},
  number={9},
  pages={1974--1984},
  year={2015},
  publisher={IEEE}
}

@article{4,
  title={PulseGAN: Learning to generate realistic pulse waveforms in remote photoplethysmography},
  author={Song, Rencheng and Chen, Huan and Cheng, Juan and Li, Chang and Liu, Yu and Chen, Xun},
  journal={IEEE Journal of Biomedical and Health Informatics},
  volume={25},
  number={5},
  pages={1373--1384},
  year={2021},
  publisher={IEEE}
}

@inproceedings{9,
  title={Lstc-rppg: Long short-term convolutional network for remote photoplethysmography},
  author={Lee, Jun Seong and Hwang, Gyutae and Ryu, Moonwook and Lee, Sang Jun},
  booktitle={Proceedings of the IEEE/CVF Conference on Computer Vision and Pattern Recognition},
  pages={6015--6023},
  year={2023}
}

@article{13,
  title={Robust remote photoplethysmography estimation with environmental noise disentanglement},
  author={Liu, Si-Qi and Yuen, Pong C},
  journal={IEEE Transactions on Image Processing},
  volume={33},
  pages={27--41},
  year={2023},
  publisher={IEEE}
}

@article{15,
  title={Dual-path tokenlearner for remote photoplethysmography-based physiological measurement with facial videos},
  author={Qian, Wei and Guo, Dan and Li, Kun and Zhang, Xiaowei and Tian, Xilan and Yang, Xun and Wang, Meng},
  journal={IEEE Transactions on Computational Social Systems},
  volume={11},
  number={3},
  pages={4465--4477},
  year={2024},
  publisher={IEEE}
}

@inproceedings{PhysDiff,
  title={Physdiff: physiology-based dynamicity disentangled diffusion model for remote physiological measurement},
  author={Qian, Wei and Su, Gaoji and Guo, Dan and Zhou, Jinxing and Li, Xiaobai and Hu, Bin and Tang, Shengeng and Wang, Meng},
  booktitle={Proceedings of the AAAI Conference on Artificial Intelligence},
  volume={39},
  number={6},
  pages={6568--6576},
  year={2025}
}

@article{rppg-toolbox,
  title={rppg-toolbox: Deep remote ppg toolbox},
  author={Liu, Xin and Narayanswamy, Girish and Paruchuri, Akshay and Zhang, Xiaoyu and Tang, Jiankai and Zhang, Yuzhe and Sengupta, Roni and Patel, Shwetak and Wang, Yuntao and McDuff, Daniel},
  journal={Advances in Neural Information Processing Systems},
  volume={36},
  pages={68485--68510},
  year={2023}
}

@inproceedings{37,
  title={VIPL-HR: A multi-modal database for pulse estimation from less-constrained face video},
  author={Niu, Xuesong and Han, Hu and Shan, Shiguang and Chen, Xilin},
  booktitle={Asian conference on computer vision},
  pages={562--576},
  year={2018},
  organization={Springer}
}

@article{38,
  title={Rhythmnet: End-to-end heart rate estimation from face via spatial-temporal representation},
  author={Niu, Xuesong and Shan, Shiguang and Han, Hu and Chen, Xilin},
  journal={IEEE Transactions on Image Processing},
  volume={29},
  pages={2409--2423},
  year={2019},
  publisher={IEEE}
}

@inproceedings{39,
  title={Image enhancement for remote photoplethysmography in a low-light environment},
  author={Xi, Lin and Chen, Weihai and Zhao, Changchen and Wu, Xingming and Wang, Jianhua},
  booktitle={2020 15th IEEE International Conference on Automatic Face and Gesture Recognition (FG 2020)},
  pages={1--7},
  year={2020},
  organization={IEEE}
}

@inproceedings{40-indoor,
  title={SparsePPG: Towards driver monitoring using camera-based vital signs estimation in near-infrared},
  author={Magdalena Nowara, Ewa and Marks, Tim K and Mansour, Hassan and Veeraraghavan, Ashok},
  booktitle={Proceedings of the IEEE conference on computer vision and pattern recognition workshops},
  pages={1272--1281},
  year={2018}
}

@article{40-drv,
  title={Near-infrared imaging photoplethysmography during driving},
  author={Nowara, Ewa M and Marks, Tim K and Mansour, Hassan and Veeraraghavan, Ashok},
  journal={IEEE transactions on intelligent transportation systems},
  volume={23},
  number={4},
  pages={3589--3600},
  year={2020},
  publisher={IEEE}
}

@article{chrom,
  title={Robust pulse rate from chrominance-based rPPG},
  author={De Haan, Gerard and Jeanne, Vincent},
  journal={IEEE transactions on biomedical engineering},
  volume={60},
  number={10},
  pages={2878--2886},
  year={2013},
  publisher={IEEE}
}

@inproceedings{LGI,
  title={Local group invariance for heart rate estimation from face videos in the wild},
  author={Pilz, Christian S and Zaunseder, Sebastian and Krajewski, Jarek and Blazek, Vladimir},
  booktitle={Proceedings of the IEEE conference on computer vision and pattern recognition workshops},
  pages={1254--1262},
  year={2018}
}

@inproceedings{DeepPhys,
  title={Deepphys: Video-based physiological measurement using convolutional attention networks},
  author={Chen, Weixuan and McDuff, Daniel},
  booktitle={Proceedings of the european conference on computer vision (ECCV)},
  pages={349--365},
  year={2018}
}

@article{TS-CAN,
  title={Multi-task temporal shift attention networks for on-device contactless vitals measurement},
  author={Liu, Xin and Fromm, Josh and Patel, Shwetak and McDuff, Daniel},
  journal={Advances in Neural Information Processing Systems},
  volume={33},
  pages={19400--19411},
  year={2020}
}

@inproceedings{PFE-TFA,
  title={Learning motion-robust remote photoplethysmography through arbitrary resolution videos},
  author={Li, Jianwei and Yu, Zitong and Shi, Jingang},
  booktitle={Proceedings of the AAAI Conference on Artificial Intelligence},
  volume={37},
  number={1},
  pages={1334--1342},
  year={2023}
}

@inproceedings{NEST,
  title={Neuron structure modeling for generalizable remote physiological measurement},
  author={Lu, Hao and Yu, Zitong and Niu, Xuesong and Chen, Ying-Cong},
  booktitle={Proceedings of the IEEE/CVF conference on computer vision and pattern recognition},
  pages={18589--18599},
  year={2023}
}

@inproceedings{Efficientphys,
  title={Efficientphys: Enabling simple, fast and accurate camera-based cardiac measurement},
  author={Liu, Xin and Hill, Brian and Jiang, Ziheng and Patel, Shwetak and McDuff, Daniel},
  booktitle={Proceedings of the IEEE/CVF winter conference on applications of computer vision},
  pages={5008--5017},
  year={2023}
}

@article{Physformer++,
  title={Physformer++: Facial video-based physiological measurement with slowfast temporal difference transformer},
  author={Yu, Zitong and Shen, Yuming and Shi, Jingang and Zhao, Hengshuang and Cui, Yawen and Zhang, Jiehua and Torr, Philip and Zhao, Guoying},
  journal={International Journal of Computer Vision},
  volume={131},
  number={6},
  pages={1307--1330},
  year={2023},
  publisher={Springer}
}

@article{Rhythmformer,
  title={Rhythmformer: Extracting patterned rppg signals based on periodic sparse attention},
  author={Zou, Bochao and Guo, Zizheng and Chen, Jiansheng and Zhuo, Junbao and Huang, Weiran and Ma, Huimin},
  journal={Pattern Recognition},
  volume={164},
  pages={111511},
  year={2025},
  publisher={Elsevier}
}

@inproceedings{Rhythmmamba,
  title={Rhythmmamba: Fast, lightweight, and accurate remote physiological measurement},
  author={Zou, Bochao and Guo, Zizheng and Hu, Xiaocheng and Ma, Huimin},
  booktitle={Proceedings of the AAAI Conference on Artificial Intelligence},
  volume={39},
  number={10},
  pages={11077--11085},
  year={2025}
}

@inproceedings{PHASE-Net,
  title={Phase-net: Physics-grounded harmonic attention system for efficient remote photoplethysmography measurement},
  author={Zhao, Bo and Guo, Dan and Cao, Junzhe and Xu, Yong and Zou, Bochao and Tan, Tao and Sun, Yue and Yu, Zitong},
  booktitle={Proceedings of the IEEE/CVF Conference on Computer Vision and Pattern Recognition},
  pages={21198--21207},
  year={2026}
}

@inproceedings{44,
  title={Physformer: Facial video-based physiological measurement with temporal difference transformer},
  author={Yu, Zitong and Shen, Yuming and Shi, Jingang and Zhao, Hengshuang and Torr, Philip HS and Zhao, Guoying},
  booktitle={Proceedings of the IEEE/CVF conference on computer vision and pattern recognition},
  pages={4186--4196},
  year={2022}
}

@inproceedings{46,
  title={Dual-gan: Joint bvp and noise modeling for remote physiological measurement},
  author={Lu, Hao and Han, Hu and Zhou, S Kevin},
  booktitle={Proceedings of the IEEE/CVF conference on computer vision and pattern recognition},
  pages={12404--12413},
  year={2021}
}

@article{55,
  title={Amplitude--time dual-view fused EEG temporal feature learning for automatic sleep staging},
  author={An, Panfeng and Zhao, Jianhui and Du, Bo and Zhao, Wenyuan and Zhang, Tingbao and Yuan, Zhiyong},
  journal={IEEE Transactions on Neural Networks and Learning Systems},
  volume={35},
  number={5},
  pages={6492--6506},
  year={2022},
  publisher={IEEE}
}

@article{vlphys,
  title={Bootstrapping vision-language models for frequency-centric self-supervised remote physiological measurement},
  author={Yue, Zijie and Shi, Miaojing and Wang, Hanli and Ding, Shuai and Chen, Qijun and Yang, Shanlin},
  journal={International Journal of Computer Vision},
  volume={133},
  number={7},
  pages={4112--4133},
  year={2025},
  publisher={Springer}
}

@inproceedings{Physllm,
  title={PhysLLM: Harnessing large language models for cross-modal remote physiological sensing},
  author={Xie, Yiping and Dai, Mingtong and Zhou, Jian-Ping and Sun, Yue and Tan, Tao and Xie, Weicheng and Shen, Linlin and YU, Zitong and others},
  booktitle={International Conference on Learning Representations},
  volume={2026},
  pages={93760--93779},
  year={2026}
}

@article{60,
  title={rPPG-MAE: Self-supervised pretraining with masked autoencoders for remote physiological measurements},
  author={Liu, Xin and Zhang, Yuting and Yu, Zitong and Lu, Hao and Yue, Huanjing and Yang, Jingyu},
  journal={IEEE Transactions on Multimedia},
  volume={26},
  pages={7278--7293},
  year={2024},
  publisher={IEEE}
}

@article{61,
  title={Self-similarity prior distillation for unsupervised remote physiological measurement},
  author={Zhang, Xinyu and Sun, Weiyu and Lu, Hao and Chen, Ying and Ge, Yun and Huang, Xiaolin and Yuan, Jie and Chen, Yingcong},
  journal={IEEE Transactions on Multimedia},
  volume={26},
  pages={10290--10305},
  year={2024},
  publisher={IEEE}
}

@article{62,
  title={HRVFusion: Video-based Long-Term Heart Rate Variability Measurement with Conditional Diffusion Models},
  author={Yao, Xing and Song, Rencheng and Cheng, Juan and Li, Chang and Chen, Xun},
  journal={IEEE Transactions on Multimedia},
  year={2026},
  publisher={IEEE}
}

@article{63,
  title={Algorithmic principles of remote PPG},
  author={Wang, Wenjin and Den Brinker, Albertus C and Stuijk, Sander and De Haan, Gerard},
  journal={IEEE Transactions on Biomedical Engineering},
  volume={64},
  number={7},
  pages={1479--1491},
  year={2016},
  publisher={IEEE}
}

@article{64,
  title={Remote photoplethysmograph signal measurement from facial videos using spatio-temporal networks},
  author={Yu, Zitong and Li, Xiaobai and Zhao, Guoying},
  journal={arXiv preprint arXiv:1905.02419},
  year={2019}
}

@inproceedings{65,
  title={Learning transferable visual models from natural language supervision},
  author={Radford, Alec and Kim, Jong Wook and Hallacy, Chris and Ramesh, Aditya and Goh, Gabriel and Agarwal, Sandhini and Sastry, Girish and Askell, Amanda and Mishkin, Pamela and Clark, Jack and others},
  booktitle={International conference on machine learning},
  pages={8748--8763},
  year={2021},
  organization={PmLR}
}

@inproceedings{66,
  title={Blip: Bootstrapping language-image pre-training for unified vision-language understanding and generation},
  author={Li, Junnan and Li, Dongxu and Xiong, Caiming and Hoi, Steven},
  booktitle={International conference on machine learning},
  pages={12888--12900},
  year={2022},
  organization={PMLR}
}

@article{67,
  title={Flamingo: a visual language model for few-shot learning},
  author={Alayrac, Jean-Baptiste and Donahue, Jeff and Luc, Pauline and Miech, Antoine and Barr, Iain and Hasson, Yana and Lenc, Karel and Mensch, Arthur and Millican, Katherine and Reynolds, Malcolm and others},
  journal={Advances in neural information processing systems},
  volume={35},
  pages={23716--23736},
  year={2022}
}

@inproceedings{68,
  title={Improved baselines with visual instruction tuning},
  author={Liu, Haotian and Li, Chunyuan and Li, Yuheng and Lee, Yong Jae},
  booktitle={Proceedings of the IEEE/CVF conference on computer vision and pattern recognition},
  pages={26296--26306},
  year={2024}
}

@article{69,
  title={Camera measurement of physiological vital signs},
  author={McDuff, Daniel},
  journal={ACM Computing Surveys},
  volume={55},
  number={9},
  pages={1--40},
  year={2023},
  publisher={ACM New York, NY}
}

@article{71,
  title={Depression recognition using remote photoplethysmography from facial videos},
  author={Casado, Constantino {\'A}lvarez and Ca{\~n}ellas, Manuel Lage and L{\'o}pez, Miguel Bordallo},
  journal={IEEE Transactions on Affective Computing},
  volume={14},
  number={4},
  pages={3305--3316},
  year={2023},
  publisher={IEEE}
}

@article{72,
  title={Mobilephys: Personalized mobile camera-based contactless physiological sensing},
  author={Liu, Xin and Wang, Yuntao and Xie, Sinan and Zhang, Xiaoyu and Ma, Zixian and McDuff, Daniel and Patel, Shwetak},
  journal={Proceedings of the ACM on Interactive, Mobile, Wearable and Ubiquitous Technologies},
  volume={6},
  number={1},
  pages={1--23},
  year={2022},
  publisher={ACM New York, NY, USA}
}

@inproceedings{flow,
  title={FLOW: Optimal Transport-Driven Feature Warping for Generalized Remote Physiological Measurement},
  author={Zhao, Bo and Cao, Junzhe and Guo, Dan and Huang, Dongmin and Wang, Wenjin and Tan, Tao and Sun, Yue and Yu, Zitong},
  booktitle={Proceedings of the IEEE/CVF Conference on Computer Vision and Pattern Recognition},
  pages={28481--28491},
  year={2026}
}

@article{qwen2,
  title={Qwen2-vl: Enhancing vision-language model's perception of the world at any resolution},
  author={Wang, Peng and Bai, Shuai and Tan, Sinan and Wang, Shijie and Fan, Zhihao and Bai, Jinze and Chen, Keqin and Liu, Xuejing and Wang, Jialin and Ge, Wenbin and others},
  journal={arXiv preprint arXiv:2409.12191},
  year={2024}
}

@inproceedings{MMPD,
  title={Mmpd: Multi-domain mobile video physiology dataset},
  author={Tang, Jiankai and Chen, Kequan and Wang, Yuntao and Shi, Yuanchun and Patel, Shwetak and McDuff, Daniel and Liu, Xin},
  booktitle={2023 45th Annual International Conference of the IEEE Engineering in Medicine \& Biology Society (EMBC)},
  pages={1--5},
  year={2023},
  organization={IEEE}
}

@article{interbvl25,
  title={Expanding performance boundaries of open-source multimodal models with model, data, and test-time scaling},
  author={Chen, Zhe and Wang, Weiyun and Cao, Yue and Liu, Yangzhou and Gao, Zhangwei and Cui, Erfei and Zhu, Jinguo and Ye, Shenglong and Tian, Hao and Liu, Zhaoyang and others},
  journal={arXiv preprint arXiv:2412.05271},
  year={2024}
}

@article{smolvlm,
  title={Smolvlm: Redefining small and efficient multimodal models},
  author={Marafioti, Andr{\'e}s and Zohar, Orr and Farr{\'e}, Miquel and Noyan, Merve and Bakouch, Elie and Cuenca, Pedro and Zakka, Cyril and Allal, Loubna Ben and Lozhkov, Anton and Tazi, Nouamane and others},
  journal={arXiv preprint arXiv:2504.05299},
  year={2025}
}

@InProceedings{VQArPPG,
    author    = {Dai, Tianyang and Chang, Ming and Chen, Yan and Hu, Yang},
    title     = {{rPPG-VQA}: A Video Quality Assessment Framework for Unsupervised {rPPG} Training},
    booktitle = {Proceedings of the IEEE/CVF Conference on Computer Vision and Pattern Recognition (CVPR)},
    month     = {June},
    year      = {2026},
    pages     = {1365--1375}
}

@article{liphysflow,
  title={PhysFlow: Frequency Decoupled with Dual-Field Rectified Flow for Remote Photoplethysmography},
  author={Li, Zixu and Shao, Hang and Luo, Lei and Yang, Jian and others},
  journal={arXiv preprint arXiv:2606.23226},
  year={2026}
}

@article{dinov3,
  title={Dinov3},
  author={Sim{\'e}oni, Oriane and Vo, Huy V and Seitzer, Maximilian and Baldassarre, Federico and Oquab, Maxime and Jose, Cijo and Khalidov, Vasil and Szafraniec, Marc and Yi, Seungeun and Ramamonjisoa, Micha{\"e}l and others},
  journal={arXiv preprint arXiv:2508.10104},
  year={2025}
}

@article{perception,
  title={Perception encoder: The best visual embeddings are not at the output of the network},
  author={Bolya, Daniel and Huang, Po-Yao and Sun, Peize and Cho, Jang Hyun and Madotto, Andrea and Wei, Chen and Ma, Tengyu and Zhi, Jiale and Rajasegaran, Jathushan and Bangalath, Hanoona and others},
  journal={Advances in Neural Information Processing Systems},
  volume={38},
  pages={60884--60937},
  year={2026}
}

@article{pei1,
  title={Video respiratory rate measurement in walking scenarios using multi-strategy adaptive denoising},
  author={Pei, Gan and Ning, Junhao and Niu, Chenrui and Yao, Siqiong and Hu, Menghan and Zhai, Guangtao},
  journal={IEEE Transactions on Circuits and Systems for Video Technology},
  year={2026},
  publisher={IEEE}
}

\vfill
\clearpage

\end{document}